%% file: arxiv.tex
\documentclass[11pt]{article}

\usepackage[preprint]{acl}
\usepackage{colortbl}
\usepackage[table]{xcolor}
\usepackage{times}
\usepackage{latexsym}
 \usepackage{booktabs}
\usepackage[T1]{fontenc}
\usepackage{multirow}

\usepackage[utf8]{inputenc}
\usepackage{amsmath} 
\usepackage{microtype}

\usepackage{inconsolata}

\usepackage{graphicx}

\usepackage{fontawesome}
\title{Plan Before Search: Search Agents Need Plan}

\author{
  \textbf{Zhipeng Qian\textsuperscript{1}\textsuperscript{*}},
  \textbf{Zihan Liang\textsuperscript{1}\textsuperscript{*}},
  \textbf{Yufei Ma\textsuperscript{1}\textsuperscript{*}},
  \textbf{Ben Chen\textsuperscript{1}\textsuperscript{†}},
  \textbf{Huangyu Dai\textsuperscript{1}},
  \textbf{Jiayi Ji\textsuperscript{2}},
\\
  \textbf{Chenyi Lei\textsuperscript{1}},
  \textbf{Wenwu Ou\textsuperscript{1}},
  \textbf{Xiaoshuai Sun\textsuperscript{2}},
  \textbf{Qibin Hou\textsuperscript{3}}
\\
\\
  \textsuperscript{1}Kuaishou Technology \\
  \textsuperscript{2}Key Laboratory of Multimedia Trusted Perception and Efficient Computing, \\
  \quad Ministry of Education of China, Xiamen University, 361005, P.R. China \\
  \textsuperscript{3}VCIP, CS, Nankai University \\
  \small{\textsuperscript{*}Equal contribution. \quad \textsuperscript{†}Corresponding author.} \\
  \small{\faEnvelope \quad \texttt{benchen4395@gmail.com}}
}

\usepackage{longtable}
\usepackage{booktabs}
\usepackage{multirow}
\usepackage{colortbl}
\usepackage{tabularray}
\UseTblrLibrary{booktabs}   

\usepackage{xcolor}
\usepackage{pifont}        
\usepackage{array}
\usepackage{tabularx}

\definecolor{drift}{HTML}{C0392B}
\definecolor{ok}{HTML}{27AE60}
\newcommand{\xmark}{\textcolor{drift}{\ding{55}}}
\newcommand{\cmark}{\textcolor{ok}{\ding{51}}}

\usepackage{xcolor}
\usepackage{pifont}
\usepackage{booktabs}
\usepackage{tabularx}

\definecolor{drift}{HTML}{C0392B}
\definecolor{ok}{HTML}{27AE60}
\definecolor{evidence}{HTML}{1F6FEB}
\definecolor{leak}{HTML}{C0392B}

\newcommand{\drifttag}{\textcolor{drift}{\textbf{[drift]}}}
\newcommand{\ev}[1]{\textcolor{evidence}{#1}}
\newcommand{\leakt}[1]{\textcolor{leak}{#1}}
\usepackage{amsfonts} 
\begin{document}
\maketitle
\begin{abstract}
Training large language models as retrieval-augmented reasoning agents typically combines reinforcement learning with an SFT cold start distilled from a stronger model. However, this paradigm overlooks two fundamental factors: the dependency structure among sub-skills, and the possibility that distillation is not the only route to capability acquisition. We study this through Plan, a structured agentic behavior for multi-hop retrieval that decomposes a question into ordered sub-questions before any retrieval is performed, so that each search step can be anchored to a pre-designed sub-question instead of drifting under the influence of partially relevant documents retrieved earlier. However,  across three model families spanning 3B to 14B parameters, we find that an identical reward signal induces qualitatively different RL failure modes. This phenomenon indicates that successful training hinges not only on reward design but also on model-specific feasibility conditions: sufficient initial entropy, training stability, and prerequisite sub-skills. Motivated by this, we propose a self-bootstrapping paradigm in which a small-scale seed model generates filtered trajectories that activate Plan in any target model, eliminating the need for distillation from an external stronger model. Our pipeline activates Plan across every tested model and consistently outperforms competitive baselines on multi-hop QA benchmarks. Our code is available at \url{https://github.com/qzp2018/PL-Search}.
\end{abstract}

\section{Introduction}

\begin{figure}[t]
    \centering
    \includegraphics[width=\linewidth]{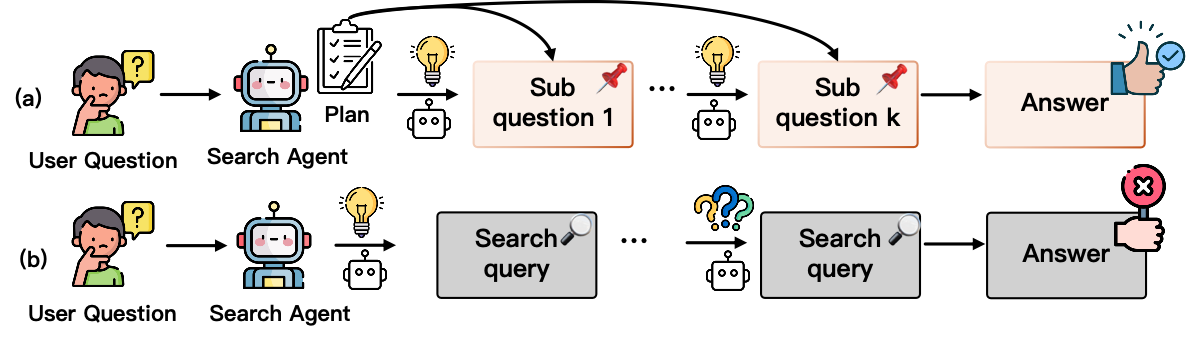}
    \caption{Comparison with traditional search agents. 
        \textbf{(a)} Our method first generates a global \emph{plan}
        that decomposes the question into ordered sub-questions; each
        subsequent search is anchored to one
        sub-question, yielding the correct answer.
        \textbf{(b)} Without a plan, the agent forms each \emph{search
        query} reactively from previously retrieved results, and the
        query progressively drifts from the original question, leading
        to a wrong answer.
    }
    \label{fig:pipeline}
\end{figure}

Recent work has demonstrated that reinforcement learning can effectively train large language models to acquire search-augmented reasoning behaviors. Search-R1~\citep{jin2025search} pioneers training models via RL to interleave thinking and retrieval within a single trajectory, using outcome-based rewards, and AutoRefine~\citep{shi2025search} extends this framework by introducing explicit document refinement and retrieval-specific rewards.
Other works, such as ReSearch~\citep{chen2026learning}, further investigate reward shaping and self-reflection mechanisms. 
When direct RL training proves insufficient for more complex behaviors, a now-standard remedy is to apply supervised fine-tuning as a cold start prior to RL optimization, typically using trajectories distilled from a stronger model. Together, careful reward design and SFT cold start with stronger-model distillation constitute the dominant toolkit for training search-augmented LLMs.

However, this paradigm overlooks two aspects of complex agentic behaviors: the role of sub-skill interdependencies in constraining trainability, and whether stronger-model distillation is truly necessary for SFT cold-start.
In this paper, we investigate both questions through the plan behavior, since the plan is particularly suited for this investigation for two reasons. 
First, it sits at the next level of complexity beyond Search-R1 and AutoRefine, requiring the coordination of multiple sub-skills that prior work has addressed individually.
Second, without an upfront decomposition of the question, each search step in prior methods largely relies on the documents retrieved in previous steps to decide what to search next. 
Noisy or partially relevant documents accumulate along the trajectory, causing the agent to drift away from the original question.
Figure~\ref{fig:pipeline} contrasts this reactive baseline (b) with our plan-based approach (a). A global plan, formulated before any retrieval is performed, anchors each subsequent reasoning step to a fixed sub-question, preventing such drift. We formalize plan as a structured agentic behavior and design a plan-aware reward function that supervises both its structural validity and the alignment between each reasoning step and its planned sub-question. Through systematic experiments across three model families (Qwen2.5, Llama3.2, and Qwen3) and scales ranging from 3B to 14B, we report two findings.

\textbf{First, direct RL training of plan fails in qualitatively different ways across models}, despite identical reward designs. Models with low initial entropy lack sufficient exploration room, and the policy cannot acquire new behaviors regardless of the reward signal. In addition, some models train stably at first but collapse abruptly. As entropy rises, output format breaks down, the reward signal vanishes, and training destabilizes irreversibly. Models that lack the refine sub-skill also fail: Plan demands that pre-formulated sub-questions remain consistent with subsequently retrieved evidence. Without the ability to integrate information across this boundary, both forcing the model to acquire plan and refine simultaneously and forcing it to learn plan first lead to training collapse. Although these failures appear different on the surface, they share a single root cause---the initial model does not satisfy the implicit conditions required for plan RL: room to explore, training stability, and the prerequisite sub-skills. A single SFT pass addresses all three conditions at once, which is why it is effective across all the tested models.

\textbf{Second, we propose a self-bootstrapping training paradigm.} Prior work constructs cold-start SFT data by distilling trajectories from a larger and more capable model. We identify an alternative and effective approach: A small seed model that satisfies the feasibility conditions identified above can acquire a plan through direct RL and generate trajectories that activate a plan in any other target model, and the self-bootstrapped data even outperform strong-model distillation in our experiments. 
Specifically, some models, exemplified by Qwen2.5-3B-Base, can autonomously acquire the plan through pure RL by first learning refinement and then learning plan on top of it. 
We exploit such a model as a \textit{seed} that generates plan trajectories, which are then filtered through a quality-aware pipeline and used as SFT data to activate the plan capability in any target model. Through this procedure, every target model, regardless of family or scale, successfully acquires plan and achieves consistent performance gains on multi-hop benchmarks, without the need for an external stronger model. We refer to the resulting method as PL-Search.
In summary, our contributions are as follows.
\begin{itemize}
    \item We formulate plan as a structured agentic behavior that prevents trajectory drift by anchoring each retrieval step to a pre-designed sub-question and name it  PL-Search, a method trained with a plan-aware reward that supervises step-level sub-question alignment.

    \item We identify three feasibility conditions for direct plan RL---sufficient initial entropy, training stability, and the presence of prerequisite sub-skills---and show that the violation of these conditions manifests as qualitatively distinct failure modes across model families.

    \item We propose a self-bootstrapping paradigm in which a small-scale seed model generates filtered trajectories to activate plan in any target model, eliminating the need for strong-model distillation while outperforming it.
\end{itemize}

\section{Related Work}
\label{sec:related}
\subsection{RL Training of Agentic LLMs}
The search field~\cite{zheng2025onevision,chen2026onesearch,liang2025uniecs,qian2026csmcir,liang2026search,liu2026bian,liang2026ig} has witnessed the rapid development of search agents.
Search-R1~\cite{jin2025search} established the paradigm of RL-trained agentic search with outcome-based rewards. Subsequent work extended this paradigm through richer reward designs and algorithmic variants, including refinement-augmented trajectories~\cite{shi2025search}, tighter integration between reasoning and retrieval~\cite{chen2026learning, song2025r1,li2026se}, step-wise credit assignment~\cite{wang2025stepsearch}, retrieval-gain optimization~\cite{jiang2025s3}, and hierarchical reasoning structures~\cite{xu2026thinker}. When direct RL proves insufficient for more complex behaviors, SFT cold-start with stronger-model distillation has become the default remedy. It is applied in some works~\cite{song2025r1, sun2025simpledeepsearcher, christianos2023pangu} but skipped in others~\cite{jin2025search, chen2026learning, shi2025search, wang2025stepsearch}, without a principled understanding of when it is necessary or why it is effective. We use plan to fill this gap: by tracing RL failures to dependencies on prerequisite sub-skills, we explain \textit{why} SFT cold-start is so consistently effective; and by showing that a small-scale seed model suffices, we demonstrate that stronger-model distillation is not a prerequisite.

\subsection{Multi-Step Retrieval-Augmented Reasoning}
Prior work on multi-step retrieval has focused primarily on pipeline design---teaching models to invoke search iteratively through SFT, or enabling adaptive query generation through RL~\cite{jin2025search, chen2026learning, zheng2025deepresearcher,wang2025stepsearch,yuan2025memsearcher,wang2026ragen,feng2026group,tian2026internvl}, with some approaches incorporating structural elements such as refinement steps~\cite{shi2025search, zhao2025parallelsearch, liu2026opera,zhang2025evolvesearch}. The question of how a single model \emph{acquires} decomposition, retrieval, and integration as jointly dependent sub-skills through RL, and what determines whether that acquisition succeeds, has not yet been addressed. Moreover, in the absence of upfront decomposition, each search query is formed reactively from previously retrieved documents, which accumulates noise and causes trajectories to drift. We study plan as a composite agentic behavior situated at this intersection, using it as a structured probe into the sub-skill dependencies that govern the acquisition of capabilities through RL.

\begin{figure*}[t]
\centering
\includegraphics[width=2\columnwidth]{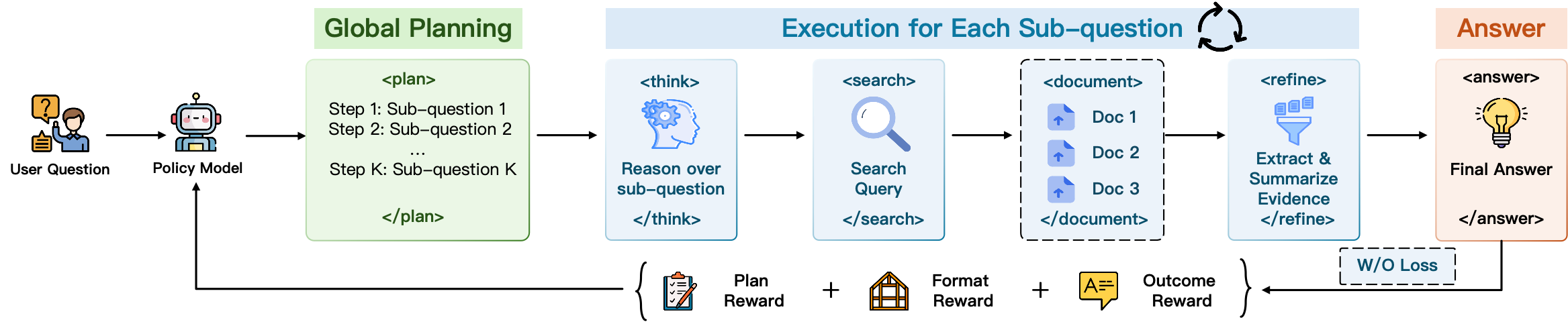}
\caption{Overview of our framework. Given a user question, the policy model first generates a global plan decomposing the question into sub-questions, then iteratively executes think, search, and refine for each sub-question before producing the final answer. Training is driven by a composite reward combining plan, format, and outcome rewards, with the loss masked over retrieved documents.}
\label{fig:overview}
\end{figure*}

\section{Methodology}
\label{sec:method}
This section presents our method for studying the prerequisite dependencies of complex agentic behaviors through plan. All RL training uses GRPO~\citep{shao2024deepseekmath} with retrieved document tokens masked from the loss, following standard practices~\citep{jin2025search, shi2025search}.

\subsection{Task Formulation of Plan}
\label{sec:plan_formulation}
As depicted in Figure~\ref{fig:overview}, we formalize plan as a structured agentic behavior that decomposes a multi-hop question into ordered sub-questions and coordinates multi-step search and refinement to integrate evidence across reasoning steps. Compared with the search-and-refine-during-think paradigm in AutoRefine, plan introduces an additional planning action that explicitly governs the decomposition of the original question and binds each subsequent reasoning step to a specific sub-question.

\paragraph{Trajectory structure.}
Given a multi-hop question $q$, the policy model $\pi_\theta$ generates a trajectory $o = (\tau_0, \tau_1, \ldots, \tau_K)$, where $\tau_0$ is a global plan and each $\tau_k$ executes one sub-question. The plan step $\tau_0$ is a single \texttt{<plan>} block listing ordered sub-questions Step 1 through Step $K$. Each execution step has the form
\begin{align}
\tau_k = \bigl(&\texttt{<think>}_k,\, \texttt{<search>}_k, \nonumber \\
&\texttt{<documents>}_k,\, \texttt{<refine>}_k\bigr),
\end{align}
where \texttt{<think>} addresses the $k$-th sub-question, \texttt{<search>} issues a search query, \texttt{<documents>} contains the retrieved documents, and \texttt{<refine>} distills them into a compact summary carried into subsequent steps. The trajectory terminates with an \texttt{<answer>} block. The full prompt template and a representative trajectory appear in Appendix~\ref{sec:example_appendix}.
\subsection{Plan-Aware Reward Design}
\label{sec:plan_reward}

The overall reward $R$ combines an outcome-based component with two auxiliary components that supervise distinct aspects of plan execution:
\begin{equation}
R = R_\text{ans} + \mathbb{I}[R_\text{ans}\!=\!0] \cdot R_\text{aux},
\end{equation}
\begin{equation}
R_\text{aux} = \lambda_\text{fmt} R_\text{fmt} + \lambda_\text{plan} R_\text{plan},
\end{equation}
where the auxiliary rewards are activated only when the answer score is zero, that is, when the predicted answer has no token overlap with the ground truth. This design ensures that auxiliary structural signals do not perturb the gradient for trajectories that already produce partially or fully correct answers.

\paragraph{Outcome reward $R_\text{ans}$.}
We compute the F1 score between the predicted answer $o_\text{ans}$ and the ground-truth answer $a$:
\begin{equation}
R_\text{ans} = \text{F1}(o_\text{ans}, a) = \frac{2|o_\text{ans} \cap a|}{|o_\text{ans}| + |a|}.
\end{equation}
The F1 formulation provides graded feedback for partially correct answers, which is more informative than exact match in early training stages.

\paragraph{Format reward $R_\text{fmt}$.}
The format reward enforces the structural validity of the trajectory. It checks two conditions: (i) all required tags (\texttt{<plan>}, \texttt{<think>}, \texttt{<search>}, \texttt{<refine>}, \texttt{<answer>}) appear with matched opening and closing markers, and (ii) every \texttt{<search>} operation is immediately preceded by a \texttt{<think>} block. The second condition is critical for plan because it enforces the principle that every retrieval action must be driven by an explicit reasoning step. We assign $R_\text{fmt} = 1$ when both conditions are satisfied, $R_\text{fmt} = 0.5$ when the first condition is met but the second is violated, and $R_\text{fmt} = 0$ otherwise.


\paragraph{Plan reward $R_\text{plan}$.}
The plan reward is the central novel component that supervises the structural coherence of plan execution. It enforces semantic alignment between the planned sub-questions and the reasoning carried out at each execution step.
Let $\{(k, p_k)\}_{k=1}^{K}$ denote the ordered sub-questions parsed from the \texttt{<plan>} block, and let $t_k$ denote the \texttt{<think>} content of the $k$-th execution step. We define the plan-think alignment score as the average token-level F1 between each pair $(p_k, t_k)$:
\begin{equation}
S_\text{align} = \frac{1}{K}\sum_{k=1}^{K} \text{F1}\big(p_k,\, t_k\big).
\end{equation}

To prevent the model from collapsing to verbatim copying of plan content into think blocks, we apply a thresholded shaping function:
\begin{equation}
R_\text{plan} = 
\begin{cases}
1, & \text{if } S_\text{align} > \delta, \\
S_\text{align}, & \text{otherwise.}
\end{cases}
\end{equation}
We set $\delta = 0.25$, which empirically marks the boundary between meaningful semantic alignment and lexical overlap noise. This thresholded design encourages the model to ground each reasoning step in the corresponding planned sub-question, without enforcing exact wording to match.

\subsection{Self-Bootstrapping Training Pipeline}
\label{sec:self_bootstrapping}

Direct RL training of plan succeeds on only a few models in our experiments, while the rest fail in heterogeneous ways (analyzed in Section~\ref{sec:analysis}). We exploit a self-bootstrapping pipeline: a seed model that satisfies the feasibility conditions (room to explore, training stability, and the prerequisite sub-skills) generates trajectories that activate plan in any target model, without requiring external distillation from a large scale model.

\paragraph{Seed model construction.}
We take Qwen2.5-3B-Base, a model that satisfies these feasibility conditions and thus acquires plan through a two-stage RL procedure. Stage 1 trains the model to acquire refine through AutoRefine-style RL~\citep{shi2025search}. Stage 2 initializes from the Stage 1 checkpoint and trains on plan under the full reward defined in Section~\ref{sec:plan_reward}. The resulting checkpoint $\pi_\text{seed}$ is then used to generate trajectories for downstream activation.

\paragraph{Trajectory generation and quality-aware filtering.}
We use $\pi_\text{seed}$ to generate plan trajectories on the training set, and then filter them through hard and soft criteria. The hard filters require each trajectory to follow the structural format \texttt{plan} $\rightarrow$ (\texttt{think} $\rightarrow$ \texttt{search} $\rightarrow$ \texttt{refine})$^+$ $\rightarrow$ \texttt{answer} and to produce an answer that matches the ground truth under cover-EM. The surviving trajectories are scored by a weighted combination of search-round count, query diversity (by bigram Jaccard distance), and refinement density. We then apply bucket sampling across search-step counts to obtain balanced coverage of reasoning depths. This procedure yields 4,500 trajectories that serve as SFT data for downstream activation. The full pipeline details are provided in Appendix~\ref{appendix:implementation}.

\paragraph{Activation via SFT and RL refinement.}
Every target model, whether base or instruction-tuned and regardless of family, follows the same procedure: SFT on the filtered seed trajectories, followed by RL refinement under the combination of plan reward, format reward and outcome reward. The SFT stage establishes the plan's structural format while calibrating policy entropy to a workable range—neither too high, risking the entropy collapse and reward hacking that crash training, nor too low, sacrificing the exploration capacity RL needs.
\input{tables/main_table}

\section{Experiments}
\subsection{Experiment Setup}
\label{sec:exp_setup}

\paragraph{Datasets and metrics.} 
Following~\citep{jin2025search, shi2025search}, we train on the combined NQ~\citep{kwiatkowski2019natural} and HotpotQA~\citep{yang2018hotpotqa} training set, and evaluate on three single-hop benchmarks (NQ, TriviaQA~\citep{joshi2017triviaqa}, PopQA~\citep{mallen2023not}) and four multi-hop benchmarks (HotpotQA, 2WikiMultihopQA~\citep{ho2020constructing}, Musique~\citep{trivedi2022musique}, Bamboogle~\citep{press2023measuring}). We report Exact Match (EM) as the primary metric; F1 and Cover Exact Match are in Appendix~\ref{appendix:detailed_results}. 



\subsection{Overall Performance}
\label{sec:overall_performance}

Table~\ref{tab_exp_main} presents the overall performance comparison between our PL-Search and baseline methods. The Avg. column denotes the average exact match (EM) accuracy. PL-Search achieves the best overall performance, with the base and instruct variants reaching 0.430 and 0.432 average EM, respectively, surpassing the strongest baseline by 0.025.
The gains are most pronounced on multi-hop QA, where PL-Search-Base improves over the strongest baseline by 0.027 on average and by 0.045 over AutoRefine-Base, isolating the contribution of the plan stage on top of refine. We attribute these gains to the model's ability to autonomously decompose complex questions and integrate evidence across reasoning steps, proving plan's effectiveness.

\begin{figure}[t]
\centering
\includegraphics[width=\linewidth]{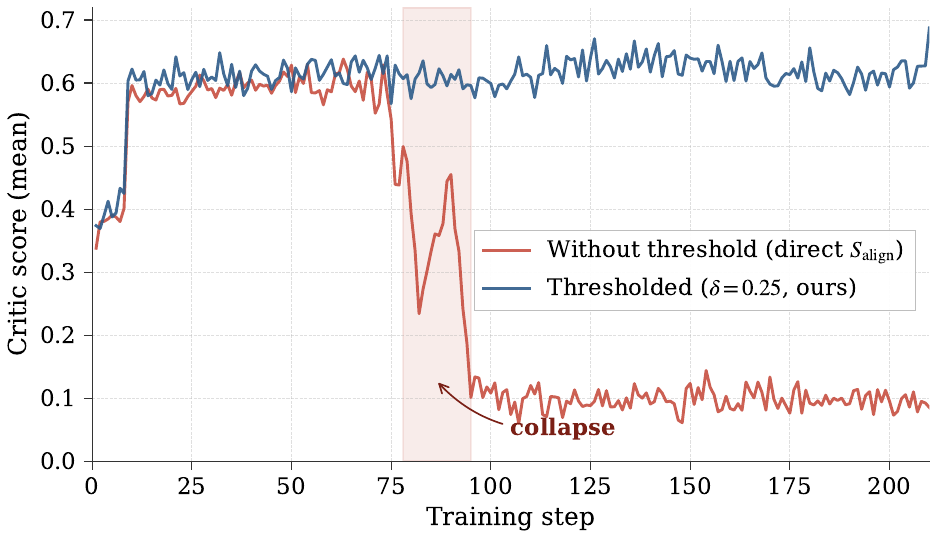}
\caption{Training stability comparison between thresholded and non-thresholded plan reward formulations on Qwen2.5-3B-Base. Without the threshold (directly using $S_\mathrm{align}$ as the reward), training collapses around step 90 as the model degenerates into verbatim copying of plan content into think blocks. The thresholded design ($\delta=0.25$) maintains stable training throughout.}
\label{fig:threshold_collapse}
\end{figure}

\input{tables/ablation_plan_align}
\input{tables/sizes_and_types}

\subsection{Ablation Study}
\label{sec:ablation}

We conduct two ablations to validate the key design choices of PL-Search: (1) the necessity of the plan reward and (2) the advantage of plan-and-refine over refine alone across model families and scales.

\paragraph{Plan reward formulation.}
We validate two design choices in $R_\text{plan}$. First, removing $R_\text{plan}$ entirely degrades MH-Avg.\ by 0.017 on Qwen2.5-3B-Base and 0.025 on Qwen2.5-3B-Instruct (Table~\ref{tab:plan_reward_ablation}), confirming it provides signal beyond the outcome reward. Second, as shown in Figure~\ref{fig:threshold_collapse} , without the threshold $\delta$, the policy collapses around step 90 by copying plan content verbatim into think blocks, crashing the critic score from 0.6 to 0.1, so $\delta=0.25$ eliminates this reward-hacking pathway.

\paragraph{Plan-and-refine vs.\ refine-only across models.}
Table~\ref{tab:main_results_em} compares PL-Search against the AutoRefine baseline (refine without explicit planning) on 12 model configurations spanning the Qwen2.5, Llama3.2, and Qwen3 families at 3B--14B scales. PL-Search yields multi-hop gains across all 12 settings, with an average MH-Avg.\ improvement of 0.053. The outlier is Llama3.2-3B-Base, where AutoRefine fails to train due to insufficient instruction-following, and our SFT enables it to refine and plan. Excluding this case, the remaining 11 settings also show considerable MH-Avg.\ improvements, proving the contribution of explicit planning. Single-hop performance is preserved across all settings, which also aligns with our motivation since the plan mainly works on multi-hop questions.

\input{tables/self_vs_distill}

\begin{figure}[t]
\centering
\includegraphics[width=\linewidth]{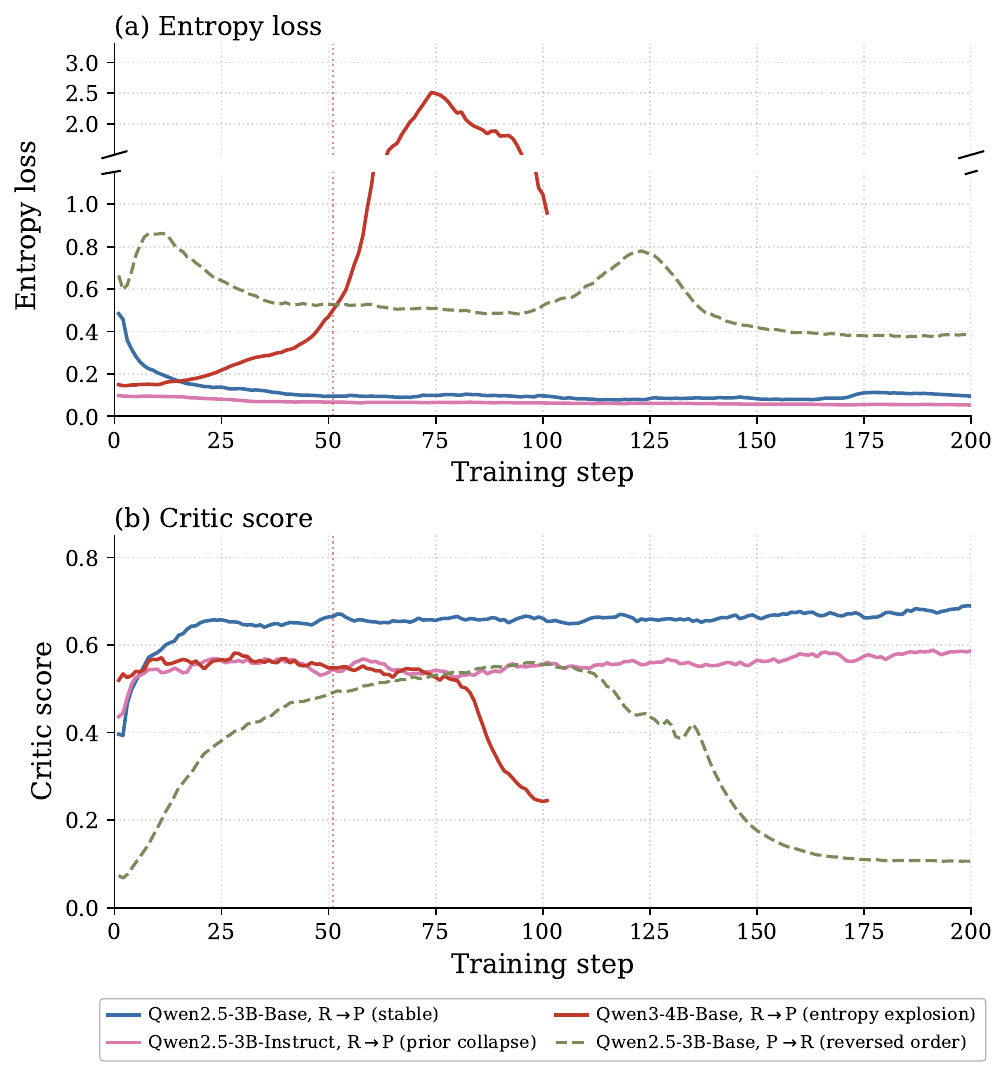}
\caption{Failure modes of direct RL on the plan task across four configurations (R: Refine, P: Plan). \textbf{(a)} Training entropy: only Qwen2.5-3B-Base under the R~$\to$~P order trains stably, whereas the other three runs exhibit prior collapse, entropy explosion, or volatile training under the reversed order. \textbf{(b)} These entropy dynamics correspond directly to downstream performance: only the stable run yields sustained EM gains, while the others either fail to improve or collapse at the step where their entropy breaks down.}
\label{fig:failure_modes}
\end{figure}

\paragraph{Self-bootstrapping vs.\ strong-model distillation.}
To examine whether the use of self-generated SFT data merely reflects a budget-constrained compromise, we replace the rollouts of the seed model with 4{,}500 correctness-filtered trajectories produced by Qwen2.5-72B-Instruct across three target models (Qwen2.5-3B-Base, Qwen2.5-7B-Base, and Qwen3-4B-Base), while holding all other components fixed. As reported in Table~\ref{tab:self_vs_distill}, self-bootstrapping consistently outperforms distillation on all three target models, yielding an average EM improvement of 1.4 points and surpassing distillation on 6 out of 7 individual benchmarks in every case; the sole exception is Bamboogle, on which the trajectories distilled from the 72B model perform better. Beyond accuracy, training stability also degrades substantially under distillation of the 72B model: for Qwen2.5-7B-Base and Qwen3-4B-Base, RL training initialized from distillation-based SFT frequently collapses, whereas initialization from self-bootstrapping remains stable throughout the entire training process. We attribute both the accuracy gap and the observed instability to distributional misalignment. The trajectories generated by the seed model lie within the reliable behavior region of the target model, so SFT produces a starting checkpoint from which RL can meaningfully proceed. In contrast, the trajectories distilled from the 72B model reflect reasoning styles and output structures that the target model cannot reliably reproduce; performing SFT on such off-support data yields a poorly aligned initialization that both degrades performance and destabilizes the subsequent RL stage. These results suggest that the quality of trajectories for a small target model depends not only on the capability of the teacher, but equally on whether the behavior of the teacher lies within the reachable distribution.

\subsection{Analysis}
\label{sec:analysis}

In this section, we examine when direct RL training of plan succeeds and when it fails. We identify three conditions required for plan RL to succeed and characterize the qualitatively distinct failure modes that arise when each is violated (\S\ref{sec:failure_modes}), then explain why SFT cold start works as a universal remedy (\S\ref{sec:sft_remedy}), motivating the self-bootstrapping pipeline in Section~\ref{sec:self_bootstrapping}.

\subsubsection{When Does Direct RL on Plan Succeed?}
\label{sec:failure_modes}

Figure~\ref{fig:failure_modes} shows entropy and critic-score dynamics across three R$\to$P (plan trained with prior refine mastery) runs on Qwen2.5-3B-Base, Qwen2.5-3B-Instruct, and Qwen3-4B-Base, plus a P$\to$R ablation  (plan trained without prior refine mastery). The results expose three conditions required for plan RL to succeed. A detail case study can be found in Appendix \ref{appendix:failure_trajectories}.

\paragraph{Condition 1:sufficient initial entropy.}
Qwen2.5-3B-Instruct exhibits \emph{prior collapse}: entropy starts near $0.06$ and remains stuck, as instruction tuning has compressed the behavioral distribution into a region too narrow to discover plan-aligned trajectories. Trajectory inspection confirms plan-formatted outputs, but with generic think blocks disengaged from the sub-questions. Qwen2.5-3B-Base, with initial entropy around $0.5$, clears this threshold.

\paragraph{Condition 2:training stability.}
Qwen3-4B-Base shows \emph{late-stage entropy explosion}: stable for about $50$ steps, then entropy surges past $2.5$. As entropy rises, output format becomes unreliable, the reward signal vanishes, and training destabilizes irreversibly---the critic score collapses. Qwen2.5-3B-Base, by contrast, decays smoothly to $0.1$ with critic score rising monotonically to $0.69$. That identical rewards succeed on one checkpoint and fail on another confirms the bottleneck is the starting point, not the reward design.

\paragraph{Condition 3:prerequisite sub-skills.}
\label{sec:training_order}
To test whether the R$\to$P order reflects a genuine dependency between sub-skills, we ablate the training order on Qwen2.5-3B-Base---the only model on which R$\to$P succeeds--- thereby isolating order effects from model-intrinsic properties. The P$\to$R curve in Figure~\ref{fig:failure_modes} exhibits volatile entropy, with peaks at $1.0$ (step 10) and $0.9$ (step 120); the critic score briefly reaches $0.55$ before collapsing to $0.10$. Without prior training on refine, each plan step must operate on raw retrieved documents; the plan reward can be satisfied superficially at first, but the lack of evidence integration destabilizes training over time. The R$\to$P order succeeds because refine provides reliable evidence integration first, lowering the exploration burden and keeping the RL gradients coherent.

\begin{figure}[t]
\centering
\includegraphics[width=\linewidth]{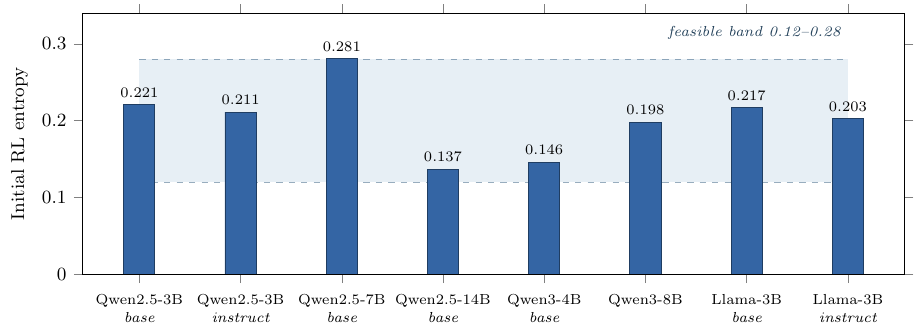}
\caption{Initial RL entropy after SFT cold start. All the models' entropy is located within a reasonable scope. }
\label{fig:post_sft_entropy}
\end{figure}

\subsubsection{Why Self-Bootstrapping Cold Start Works as a Universal Remedy}
\label{sec:sft_remedy}
The heterogeneous failures in \S\ref{sec:failure_modes} differ in symptom but share a common structural cause: the initial policy occupies a region of the policy space from which RL cannot drive sustained improvement. This region is either too narrow (prior collapse), too unstable (entropy explosion), or too deficient in prerequisite capability (P$\to$R volatility). A single SFT pass on in-distribution trajectories resolves all three problems simultaneously: it instills the structural format of plan, broadens the behavioral prior of instruction-tuned models, and implicitly transfers the capability of refine. As Figure~\ref{fig:post_sft_entropy} shows, the post-SFT entropy consistently falls within $0.12$--$0.28$ across all eight runs spanning the Qwen2.5, Qwen3, and Llama families at 3B--14B, well within the band that supports stable RL.
Critically, this relocation requires trajectories that are not only structurally correct but also distributionally aligned with the target model. Table~\ref{tab:self_vs_distill} shows that SFT on 72B-distilled trajectories can still destabilize subsequent RL despite being structurally valid, because those trajectories lie outside the reachable distribution of the target model. Self-bootstrapping satisfies both conditions by construction: a small seed model produces trajectories that the target model can reliably reproduce; however, simply distilling a strong model fails to do so.
\section{Conclusion}
\label{sec:conclusion}

We study the dependency structure among sub-skills in RL training of search-augmented LLMs, using plan as a research vehicle. Through experiments across three model families and scales from 3B to 14B, we identify two findings. First, direct RL training of plan exhibits heterogeneous failure modes across models, revealing that the bottleneck of RL-driven acquisition lies in sub-skill dependencies rather than reward design. Second, a self-bootstrapping training paradigm, in which a single seed model generates plan trajectories to activate plan capability in other models of comparable scale, can eliminate the reliance on external strong-model distillation that current cold-start practices depend on. We hope these findings motivate further investigation into the structural properties of multi-skill agentic behaviors in retrieval and related settings.

\section*{Limitations}
\label{sec:limitations}

Our study has two main limitations. First, the investigation focuses on plan as a representative composite agentic behavior in multi-hop question answering; extending the analysis to other agentic tasks, such as tool selection and long-horizon planning, would help establish how broadly the observed dependency phenomena generalize. Second, following prior work~\citep{jin2025search, shi2025search}, our experiments use a static Wikipedia snapshot rather than a live web retrieval environment, which leaves the behavior of the proposed pipeline under dynamic and adversarial retrieval conditions to be examined in future work.
\bibliography{custom}

\appendix
\clearpage

\section{Implementation Details}
\label{appendix:implementation}

\paragraph{Retrieval setup.} We use the December 2018 Wikipedia dump~\citep{karpukhin2020dense} as the external knowledge corpus and E5-base-v2~\citep{wang2022text} as the retriever, consistent with Search-R1 and AutoRefine. The retriever returns the top three most relevant documents for each query.

\paragraph{Seed model construction.} The seed model is obtained by applying a two-stage RL procedure to Qwen2.5-3B-Base. Stage 1 trains the model with AutoRefine-style RL for 200 steps. Stage 2 initializes from the Stage 1 checkpoint and trains the model on the plan task using our full plan-aware reward function for an additional 200 steps. The resulting checkpoint is used solely as a trajectory generator for the self-bootstrapping pipeline.

\paragraph{Self-bootstrapping pipeline.} For every target model reported in our main results, we apply the same self-bootstrapping procedure: we generate plan trajectories from the seed model on the training set, filter them through the quality-aware pipeline described below to obtain 4,500 high-quality samples, perform SFT on the target model for 3 epochs, and then apply RL refinement under our plan-aware reward function for 200 steps.

\paragraph{Trajectory filtering pipeline.} The filtering pipeline processes seed-generated trajectories through four stages: (i) hard filter on format validity (enforcing \texttt{plan} $\rightarrow$ (\texttt{think} $\rightarrow$ \texttt{search} $\rightarrow$ \texttt{refine})$^+$ $\rightarrow$ \texttt{answer} structure), (ii) hard filter on answer correctness via cover-EM against the ground-truth answer, (iii) soft scoring via a quality score in $[0,1]$ that linearly combines three normalized signals: \emph{search rounds} (weight 0.40), a log-scaled count of \texttt{search} calls that saturates at four rounds; \emph{query diversity} (0.35), the mean pairwise bigram Jaccard distance across search queries, penalizing repetitive searches; and \emph{refine density} (0.25), the fraction of the response occupied by \texttt{refine} content, and (iv) bucket sampling by search step count, with target ratios of 20\%/50\%/20\%/10\% for 1/2/3/4+-step trajectories. The bucket sampling ensures the SFT data covers a balanced distribution of reasoning depths.

\paragraph{Hyperparameters and Infrastructure.} Table~\ref{tab:hyperparams} summarizes the key hyperparameters used in our experiments. All experiments are conducted on 8$\times$NVIDIA A100-80GB GPUs. We use the VeRL framework~\citep{sheng2025hybridflow} for distributed training.

\begin{table}[t]
\centering
\small
\begin{tabular}{ll}
\toprule
Parameter & Value \\
\midrule
\multicolumn{2}{l}{\textit{RL training (GRPO)}} \\
Learning rate & $1 \times 10^{-6}$ \\
Batch size & 256 \\
Micro batch size & 64 \\
Max response length & 2048 \\
Max search actions & 5 \\
Group size $G$ & 5 \\
KL coefficient $\beta$ & 0.001 \\
Clip ratio $\epsilon$ & 0.2 \\
Rollout temperature & 1.0 \\
Training steps (seed) & 200 per stage \\
Training steps (target) & 200 \\
\midrule
\multicolumn{2}{l}{\textit{Reward coefficients}} \\
$\lambda_\text{fmt}$ & 0.1 \\
$\lambda_\text{plan}$ & 0.1 \\
Plan reward threshold $\delta$ & 0.25 \\
\midrule
\multicolumn{2}{l}{\textit{SFT (for self-bootstrapping)}} \\
Learning rate & $5 \times 10^{-6}$ \\
Schedule & Cosine \\
Batch size & 64 \\
Epochs & 3 \\
Filtered samples & 4,500 \\
\bottomrule
\end{tabular}
\caption{Hyperparameters for RL training and SFT.}
\label{tab:hyperparams}
\end{table}

\section{More Ablation Studies}

\begin{figure}[t]
    \centering
    \includegraphics[width=\linewidth]{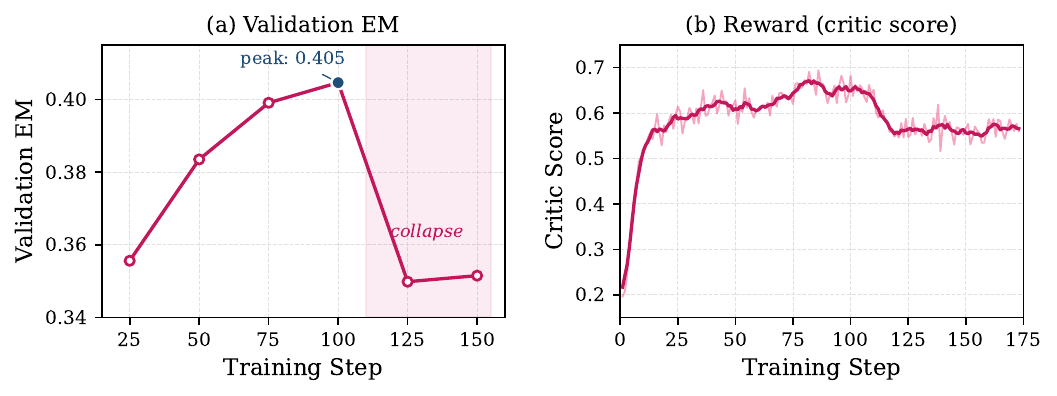}
    \caption{Training dynamics of direct joint RL on Qwen2.5-7B-Base. (a) Validation EM rises to $0.405$ at step 100, then collapses to 0.35 and fails to recover. (b) The critic score declines over the same window, suggesting the reward signal detects the collapse.}
    \label{fig:direct_rl_collapse}
\end{figure}
\subsection{Self-Bootstrapping vs.\ Direct RL on Qwen2.5-7B-Base}
\label{appendix:7b_comparison}
Qwen2.5-7B-Base is an intermediate case: unlike smaller models that collapse immediately under direct joint RL, a fraction of its trajectories acquires a partially correct format and receives non-zero reward, creating the appearance of progress. However, the majority remain malformed with zero reward, making the gradient signal too sparse to sustain training. As shown in Figure~\ref{fig:direct_rl_collapse}, validation EM peaks near $0.405$ around step 100 before collapsing, with the critic score detecting the degradation; Table~\ref{tab:7b_comparison} reports the best pre-collapse checkpoint.
Self-bootstrapping outperforms this unstable baseline by $+0.056$ EM on average ($+0.066$ on multi-hop, $+0.043$ on single-hop), by establishing a reliable format prior before RL so that reward signal is consistently accessible throughout training. The value of self-bootstrapping therefore extends beyond enabling otherwise-infeasible training: even when direct RL appears viable, it provides a fundamentally more stable optimization path.

\input{tables/appendix_7b_rl}
\input{tables/sft_ablation}

\subsection{Scaling the self-bootstrapped SFT data.}
We vary the number of correctness-filtered seed-model trajectories used for SFT from 1{,}500 to 4{,}500, keeping all other components fixed. As shown in Table~\ref{tab:sft_scale}, final RL performance improves steadily but modestly with scale (average EM $0.423 \rightarrow 0.429 \rightarrow 0.430$). The gains concentrate on the multi-hop benchmarks ($0.360 \rightarrow 0.370$), while single-hop performance saturates almost immediately. This suggests the prerequisite-acquisition stage requires only a modest amount of self-generated data to configure a usable RL starting point, and that additional data primarily sharpens the harder multi-hop behaviors rather than uniformly lifting all tasks. So we choose 4500 as our default SFT data number.

\input{tables/plan_threshold}
\subsection{Ablation on the Plan-Triggering Threshold}
\label{appendix:plan_threshold}
The plan reward threshold $\delta$ controls the boundary of the $R_\text{plan}$ reward shaping: when the plan--think alignment score $S_\text{align}$ exceeds $\delta$, the model receives full reward ($R_\text{plan}=1$); otherwise it receives the raw alignment score ($R_\text{plan}=S_\text{align}$). To assess its sensitivity, we compare our default setting ($\delta=0.25$) against fixed thresholds of $0.5$ and $0.75$, holding the base model, SFT-then-RL pipeline, and RL recipe constant. As shown in Table~\ref{tab:plan_threshold}, the default threshold attains the best average performance, outperforming the $0.5$ and $0.75$ variants by $0.010$ and $0.014$ on EM, respectively, with consistent per-dataset gains on 6 of 7 benchmarks; the only exception is Bamboogle, whose small evaluation set makes per-dataset numbers noisier. The improvement is most pronounced on the multi-hop benchmarks, consistent with the role of plan behavior in coordinating multi-step reasoning. Overall, the performance proves that 0.25 is a suitable choice of threshold.

\section{Example Appendix}
\label{sec:example_appendix}
\begin{figure}[t] 
  \centering 
  \includegraphics[width=\columnwidth]{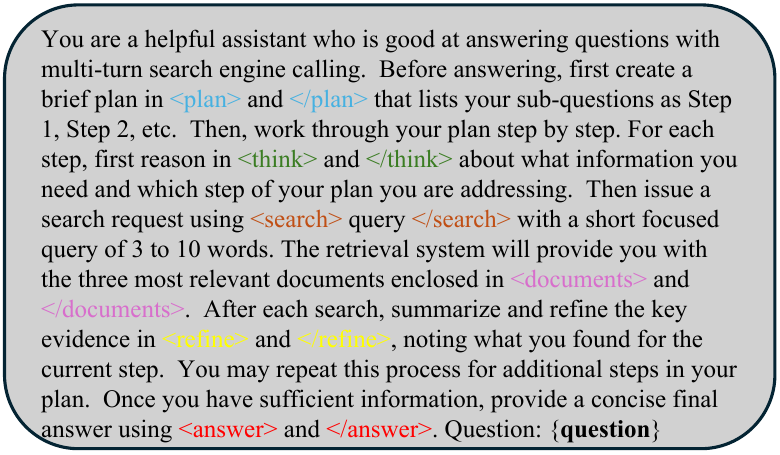} 
  \vspace{-0.8cm}
  \caption{Prompt template for plan}
  \vspace{-0.4cm}
  \label{fig:prompt_template}
\end{figure}
\subsection{Prompt Template}
\label{appendix:prompt_template}
Figure~\ref{fig:prompt_template} shows the prompt template used in our method. The template enforces a two-phase structure. In the \emph{planning phase}, the model produces a \texttt{<plan>} block listing all ordered sub-questions before any retrieval, preventing later search steps from drifting under partially relevant documents. In the \emph{execution phase}, the model iterates over each sub-question through a \texttt{<think>}\,$\rightarrow$\,\texttt{<search>}\,$\rightarrow$\,\texttt{<documents>}\,$\rightarrow$\,\texttt{<refine>} cycle: \texttt{<think>} grounds reasoning in the planned sub-question, \texttt{<search>} issues a focused query, \texttt{<documents>} receives retrieved content (masked from the training loss), and \texttt{<refine>} distills evidence for subsequent steps. Finally, the trajectory ends with \texttt{<answer>}.


\subsection{Token Budget Analysis}
\label{app:token_budget}

\begin{figure}[t]
    \centering
    \includegraphics[width=\linewidth]{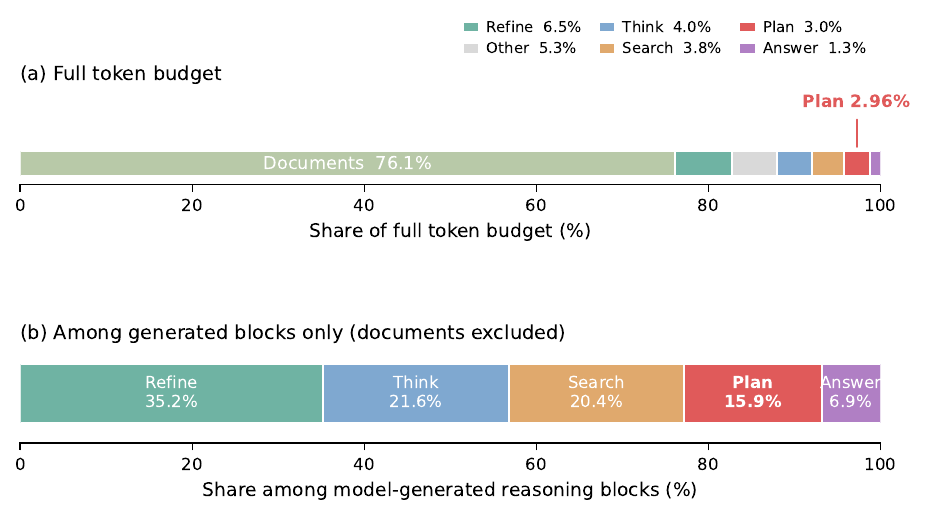}
    \caption{Per-component token distribution across training rollouts.
\texttt{Plan} takes only 2.96\% of the full budget (a) and 15.9\% among
generated blocks (b), adding negligible overhead.}
    \label{fig:token_budget}
\end{figure}

To address the concern that the multi-hop gains reported in
Table~\ref{tab:main_results_em} might stem from a larger generation budget rather than from explicit planning, we analyze the per-component token distribution across all training rollouts. Token counts are computed with the model's native tokenizer over the full rollout corpus, and the retrieved \texttt{<documents>} spans are reported separately since they are inserted by the retrieval system rather than generated by the
policy. As shown in Figure~\ref{fig:token_budget}, retrieved documents account for the overwhelming majority of the token budget (76.1\%), whereas the \texttt{<plan>} block adds only 2.96\%. Even when documents are excluded and we consider only model-generated reasoning blocks, planning remains one of the lightest components (15.9\%), below refine, think, and search. The planning action therefore introduces negligible generation overhead, and the observed improvements cannot be attributed to a substantially longer generation budget.

\subsection{Successful Trajectory Example}
\label{appendix:trajectory_example}
\input{tables/success_trajectory}
Table~\ref{tab:trajectory_case} presents a representative trajectory generated by our method on a complex multi-hop question. The example illustrates how plan decomposes the original question into three ordered sub-questions, and how each subsequent execution step grounds its reasoning, retrieval, and refinement in the corresponding planned sub-question. Notably, the model commits to a complete decomposition (Step 1: identify the team leader; Step 2: determine the ship name; Step 3: identify the designer) before any retrieval is performed, and each \texttt{<think>} block explicitly references the sub-question it addresses. The refinement steps progressively integrate evidence across hops, ultimately leading to the correct answer.

\section{Qualitative Case Studies}
\label{app:cases}

\input{tables/case_freud}
Table~\ref{tab:case_ballondor} illustrates how plan-based decomposition prevents \emph{cross-hop query drift}. The question is two-hop: it asks for the 2007 Ballon d'Or winner, then for that player's 2012 club. Without an explicit decomposition step, the reactive baseline collapses both hops into a single query that mixes the \emph{2007 award} with the \emph{2012 club}. This query targets neither sub-question cleanly, and retrieval drifts to adjacent award years (the 2011 and 2012 Ballon d'Or pages), binding the answer to that year's winner, Messi. The plan avoids this structurally: Step~1 is scoped only to resolving the 2007 winner (\emph{Kaká}), and Step~2's query is anchored to that resolved entity, retrieving Kaká's 2012 club. This is the core mechanism of plan: decomposing a compound question into ordered atomic sub-goals keeps each search query tied to a single pre-specified sub-question, preventing the cross-hop contamination that drives the reactive baseline off-target.


\subsection{Representative Failure Trajectories}
\label{appendix:failure_trajectories}
To complement the quantitative analysis in Section~\ref{sec:failure_modes}, Table~\ref{tab:failure_cases} presents three representative trajectories from direct RL training on the plan task, one per feasibility condition. Case~1 is drawn from Qwen2.5-3B-Instruct (Condition~1: prior collapse); Case~2 from Qwen3 (Condition~2: late-stage entropy explosion); and Case~3 from Qwen2.5-3B-Base trained directly on plan without first mastering refine (P$\to$R order; Condition~3: missing prerequisite sub-skills). We highlight problematic regions in \textcolor{red}{red} and key evidence, when present, in \textcolor{blue}{blue}.
Each condition produces a qualitatively distinct breakdown. Prior collapse (Case~1) yields structurally compliant outputs whose \texttt{<think>} blocks are semantically disconnected from the planned sub-questions: the model follows the format but never grounds its reasoning in the retrieved evidence. Late-stage entropy explosion (Case~2) manifests as structural drift and lexical breakdown once the policy destabilizes. Missing prerequisite sub-skills (Case~3) produces superficial format compliance at first, but the absence of reliable evidence integration destabilizes the policy and eventually causes training collapse. Despite identical reward designs and hyperparameters, these trajectories diverge in ways that no single reward signal can correct, confirming that the bottleneck is model-specific training feasibility rather than reward design.



\input{tables/fail_trajectory}

\section{Detailed Performance Results}
\label{appendix:detailed_results}
Table~\ref{tab:detailed_results} reports the per-dataset EM, F1, and CEM scores for all models and methods. The results are grouped by model family (Qwen2.5, Llama3.2, Qwen3) and by scale, comparing AutoRefine against the proposed method across three single-hop benchmarks (NQ, TriviaQA, PopQA) and four multi-hop benchmarks (HotpotQA, 2WikiMultiHopQA, MuSiQue, Bamboogle).
Two patterns are consistent across all settings. First, the gains are largest on the multi-hop benchmarks (MH-Avg.\ $\gg$ SH-Avg.), which confirms that the benefit arises from explicit decomposition rather than from general gains in capacity. Second, the improvements hold across all three model families and scales (3B--14B). One result warrants caution: the unusually large gain on Llama3.2-3B-Base (Avg.\ EM: $0.234 \to 0.408$) should not be interpreted as evidence for the particular effectiveness of plan on weak backbones. Instead, the low AutoRefine baseline on this model largely reflects poor instruction-following, which prevents the model from adhering to the search-and-refine format and therefore renders the gap a confound rather than a clean signal.

\onecolumn                          

\input{tables/sizes_and_types_appendix}          
\twocolumn                          
\end{document}

%% file: tables/main_table.tex
\begin{table*}[!ht]
    \centering
    \resizebox{\linewidth}{!}{
    \begin{tabular}{lcccccccccc}
        \toprule
        & \multicolumn{4}{c}{Single-Hop QA } & \multicolumn{5}{c}{Multi-Hop QA }       & \multicolumn{1}{l}{QA } \\
        \cmidrule(lr){2-5} \cmidrule(lr){6-10} \cmidrule(lr){11-11}
        Methods            & NQ      & TriviaQA   & PopQA  & SH-Avg.   & HotpotQA  & 2Wiki & Musique & Bamboogle & MH-Avg.  & Avg. \\
        \midrule
        \multicolumn{9}{l}{{w/o Retrieval}} \\
        \quad Direct Generation     & 0.106 & 0.288 & 0.108 & 0.167          & 0.149 & 0.244 & 0.020 & 0.024 & 0.109       & 0.134 \\
        \quad R1-Instruct \cite{guo2025deepseek}          & 0.210   & 0.449      & 0.171  & 0.277 & 0.208    & 0.275 & 0.060   & 0.192  & 0.184   & 0.224 \\
        \quad R1-Base \cite{guo2025deepseek}              & 0.226   & 0.455      & 0.173 & 0.285  & 0.201    & 0.268 & 0.055   & 0.224 &0.187     & 0.229 \\
        \midrule
        \multicolumn{9}{l}{{Workflow w/ Retrieval}} \\
        \quad Naive RAG \cite{lewis2020retrieval}                  & 0.348   & 0.544      & 0.387 & 0.426  & 0.255    & 0.226 & 0.047   & 0.080   & 0.152  & 0.270 \\
        \quad IRCoT \cite{trivedi2023interleaving} & 0.111   & 0.312      & 0.200  & 0.208 & 0.164    & 0.171 & 0.067   & 0.240  & 0.161   & 0.181 \\
        \midrule
        \multicolumn{9}{l}{{Agent w/ Retrieval}} \\
        \quad Search-o1 \cite{li2025search}                 & 0.238   & 0.472      & 0.262 & 0.324  & 0.221    & 0.218 & 0.054   & 0.320   & 0.203  & 0.255 \\
        \quad Search-R1-Instruct \cite{jin2025search}       & 0.397   & 0.565      & 0.391 & 0.451  & 0.331    & 0.310 & 0.124   & 0.232  & 0.249   & 0.336 \\
        \quad Search-R1-Base \cite{jin2025search}           & 0.421   & 0.583      & 0.413  & 0.472 & 0.297    & 0.274 & 0.066   & 0.128  & 0.191   & 0.312 \\
        \quad ReSearch-Instruct \cite{chen2026learning}         & 0.365                & 0.571                & 0.395 & 0.444 & 0.351 & 0.272 & 0.095 & 0.266 & 0.246 & 0.331 \\
        \quad ReSearch-Base \cite{chen2026learning}             & 0.427          & 0.597          & 0.430 & 0.485 & 0.305 & 0.272 & 0.074 & 0.128 & 0.195 & 0.319 \\
        \quad ZeroSearch-Base \cite{sun2025zerosearch}             & 0.430          & 0.616          & 0.414 & 0.487 & 0.338 & 0.346 & 0.130 & 0.139 & 0.238 & 0.345 \\
        \quad StepSearch-Base \cite{wang2025stepsearch}     & -    & -          & - & - & 0.329 & 0.339 & 0.181 & 0.328 & 0.294 & - \\
        \quad $O^2$-Searcher
        \cite{mei20252}       & 0.444   & 0.597      & 0.429  & 0.490 & 0.388    & 0.374 & 0.160   & 0.344 & 0.317    & 0.391 \\
        \quad AutoRefine-Instruct \cite{shi2025search}             & 0.436 & 0.597 & 0.447 & 0.493 & 0.404 & 0.380 & 0.169 & 0.336 & 0.322 & 0.396 \\
        \quad AutoRefine-Base \cite{shi2025search}                   & 0.467 & 0.620 & 0.450 & 0.512  & 0.405 & 0.393 & 0.157 & 0.344 & 0.325 & 0.405 \\        
        \quad InForage
        \cite{qian2026scent}       & 0.421   & 0.597      & 0.452 & 0.490  & 0.409    & 0.428 & 0.172   & 0.360   & 0.342  & 0.405 \\
        \quad CriticSearch \cite{zhang2025criticsearch}       & -    & -          & - & - & 0.414 & 0.409 & 0.180 & 0.368 & 0.343 & - \\
        \midrule
        \quad PL-Search-3B-Base (SFT only)        & 0.382 & 0.567 & 0.400 & 0.450 & 0.363 & 0.364 & 0.176 & 0.403 & 0.326 & 0.379 \\
\rowcolor{gray!25}
\quad \textbf{PL-Search-3B-Base} (Ours) & 0.455 & 0.623 & {0.452} & 0.510 & {0.434} & \textbf{0.429} & \textbf{0.212} & 0.403 & \textbf{0.370} & 0.430 \\
\rowcolor{gray!25}
\quad \textbf{PL-Search-3B-Instruct} (Ours) & \textbf{0.462} & \textbf{0.631} & \textbf{0.460} & \textbf{0.518} & \textbf{0.437} & 0.422 & 0.207 & \textbf{0.408} & 0.369 & \textbf{0.432} \\
        \bottomrule
        \end{tabular}
    }
    \caption{
  Main results on seven open-domain QA benchmarks (Exact Match).
  All retrieval-augmented baselines use 3B-parameter LLMs for fair
  comparison. PL-Search-Base/Instruct apply our self-bootstrapping
  pipeline (SFT on filtered seed trajectories followed by plan-aware
  RL refinement) to Qwen2.5-3B-Base/Instruct; the \emph{SFT only}
  row reports the same target after SFT but before RL refinement.
  \textbf{Bold} indicates the best result in each column.
  Avg.\ columns report per-category and overall averages.
}
    \label{tab_exp_main}
\end{table*}

%% file: tables/ablation_plan_align.tex
\begin{table}[t]
\centering
\small
\resizebox{0.49\textwidth}{!}{
\begin{tabular}{llccccc}
\toprule
Model & Method & HotpotQA & 2Wiki & Musique & Bamb. & MH-Avg. \\
\midrule
\multirow{2}{*}{Qwen2.5-3B-Base}
& Full & 0.434 & 0.429 & 0.212 & 0.403 & 0.370 \\
& w/o $R_\text{plan}$ & 0.419 & 0.419 & 0.196 & 0.376 & 0.353 ($-0.017$) \\
\midrule
\multirow{2}{*}{Qwen2.5-3B-Instruct}
& Full & 0.437 & 0.422 & 0.207 & 0.408 & 0.369 \\
& w/o $R_\text{plan}$ & 0.413 & 0.415 & 0.180 & 0.368 & 0.344 ($-0.025$) \\
\bottomrule
\end{tabular}}
\caption{Ablation study on $R_\text{plan}$, removing it consistently degrades multi-hop performance.}
\label{tab:plan_reward_ablation}
\end{table}

%% file: tables/sizes_and_types.tex
\begin{table*}[t]
    \centering
    \small
    \resizebox{\linewidth}{!}{
    \setlength{\tabcolsep}{4pt}
    \begin{tabular}{llcccccccccc}
        \toprule
        & & \multicolumn{4}{c}{Single-Hop QA} & \multicolumn{5}{c}{Multi-Hop QA} & \\
        \cmidrule(lr){3-6} \cmidrule(lr){7-11}
        Model & Method & NQ & TriviaQA & PopQA & SH-Avg. & HotpotQA & 2Wiki & Musique & Bamb. & MH-Avg. & Avg. \\
        \midrule
        \rowcolor[gray]{0.9}
        \multicolumn{12}{l}{\textit{Qwen2.5 Family}} \\
        \multirow{2}{*}{Qwen2.5-3B-Base}
            & AutoRefine & \textbf{0.467} & 0.620 & 0.450 & \textbf{0.512} & 0.405 & 0.393 & 0.157 & 0.344 & 0.325 & 0.405 \\
            & Ours       & 0.455 & \textbf{0.623} & \textbf{0.452} & 0.510 & \textbf{0.434} & \textbf{0.429} & \textbf{0.212} & \textbf{0.403} & \textbf{0.370} & \textbf{0.430} \\
        \cmidrule(lr){2-12}
        \multirow{2}{*}{Qwen2.5-3B-Instruct}
            & AutoRefine & 0.436 & 0.597 & 0.447 & 0.493 & 0.404 & 0.380 & 0.169 & 0.336 & 0.322 & 0.396 \\
            & Ours       & \textbf{0.462} & \textbf{0.631} & \textbf{0.460} & \textbf{0.518} & \textbf{0.437} & \textbf{0.422} & \textbf{0.207} & \textbf{0.408} & \textbf{0.369} & \textbf{0.432} \\
        \cmidrule(lr){2-12}
        \multirow{2}{*}{Qwen2.5-7B-Base}
            & AutoRefine & 0.484 & 0.659 & \textbf{0.487} & \textbf{0.543} & 0.451 & 0.405 & 0.187 & \textbf{0.512} & 0.389 & 0.455 \\
            & Ours       & \textbf{0.490} & \textbf{0.665} & 0.468 & 0.541 & \textbf{0.464} & \textbf{0.447} & \textbf{0.227} & 0.464 & \textbf{0.401} & \textbf{0.461} \\
        \cmidrule(lr){2-12}
\multirow{2}{*}{Qwen2.5-7B-Instruct}
    & AutoRefine & 0.423 & 0.627 & 0.430 & 0.493 & 0.409 & 0.348 & 0.175 & 0.440 & 0.343 & 0.407 \\
    & Ours       & \textbf{0.467} & \textbf{0.641} & \textbf{0.461} & \textbf{0.523} & \textbf{0.452} & \textbf{0.447} & \textbf{0.231} & \textbf{0.468} & \textbf{0.400} & \textbf{0.452} \\
        \cmidrule(lr){2-12}
        \multirow{2}{*}{Qwen2.5-14B-Base}
            & AutoRefine & \textbf{0.516} & 0.693 & \textbf{0.489} & \textbf{0.566} & 0.468 & 0.468 & 0.230 & \textbf{0.536} & 0.426 & 0.486 \\
            & Ours       & 0.514 & \textbf{0.694} & 0.487 & 0.565 & \textbf{0.500} & \textbf{0.484} & \textbf{0.269} & 0.512 & \textbf{0.441} & \textbf{0.494} \\
        \cmidrule(lr){2-12}
        \multirow{2}{*}{Qwen2.5-14B-Instruct}
            & AutoRefine & 0.483 & 0.673 & 0.451 & 0.535 & 0.463 & 0.423 & 0.231 & 0.512 & 0.407 & 0.462 \\
            & Ours       & \textbf{0.510} & \textbf{0.694} & \textbf{0.483} & \textbf{0.563} & \textbf{0.499} & \textbf{0.493} & \textbf{0.260} & \textbf{0.528} & \textbf{0.445} & \textbf{0.496} \\
        \midrule
        \rowcolor[gray]{0.9}
        \multicolumn{12}{l}{\textit{Llama3.2 Family}} \\
        \multirow{2}{*}{Llama3.2-3B-Base}
            & AutoRefine & 0.267 & 0.488 & 0.280 & 0.345 & 0.239 & 0.248 & 0.030 & 0.087 & 0.151 & 0.234 \\
            & Ours       & \textbf{0.443} & \textbf{0.612} & \textbf{0.450} & \textbf{0.502} & \textbf{0.411} & \textbf{0.388} & \textbf{0.180} & \textbf{0.368} & \textbf{0.337} & \textbf{0.408} \\
        \cmidrule(lr){2-12}
        \multirow{2}{*}{Llama3.2-3B-Instruct}
            & AutoRefine & 0.437 & \textbf{0.638} & 0.419 & 0.498 & 0.375 & 0.225 & 0.169 & 0.349 & 0.280 & 0.373 \\
            & Ours       & \textbf{0.478} & 0.626 & \textbf{0.459} & \textbf{0.521} & \textbf{0.443} & \textbf{0.390} & \textbf{0.197} & \textbf{0.468} & \textbf{0.375} & \textbf{0.437} \\
        \midrule
        \rowcolor[gray]{0.9}
        \multicolumn{12}{l}{\textit{Qwen3 Family}} \\
        \multirow{2}{*}{Qwen3-4B-Base}
            & AutoRefine & 0.418 & 0.642 & \textbf{0.500} & 0.520 & 0.418 & 0.361 & 0.157 & 0.464 & 0.350 & 0.423 \\
            & Ours       & \textbf{0.473} & \textbf{0.651} & 0.466 & \textbf{0.530} & \textbf{0.456} & \textbf{0.428} & \textbf{0.223} & \textbf{0.476} & \textbf{0.396} & \textbf{0.453} \\
        \cmidrule(lr){2-12}
        \multirow{2}{*}{Qwen3-4B-Instruct}
            & AutoRefine & 0.435 & \textbf{0.633} & 0.443 & \textbf{0.504} & 0.398 & 0.368 & 0.205 & 0.460 & 0.358 & 0.420 \\
            & Ours       & \textbf{0.437} & 0.624 & \textbf{0.448} & 0.503 & \textbf{0.437} & \textbf{0.448} & \textbf{0.217} & \textbf{0.488} & \textbf{0.398} & \textbf{0.443} \\
        \cmidrule(lr){2-12}
        \multirow{2}{*}{Qwen3-8B}
            & AutoRefine & 0.450 & \textbf{0.665} & 0.470 & 0.528 & 0.421 & 0.369 & 0.219 & \textbf{0.496} & 0.376 & 0.441 \\
            & Ours       & \textbf{0.454} & 0.660 & \textbf{0.476} & \textbf{0.530} & \textbf{0.461} & \textbf{0.453} & \textbf{0.233} & 0.475 & \textbf{0.406} & \textbf{0.459} \\
        \cmidrule(lr){2-12}
        \multirow{2}{*}{Qwen3-14B}
            & AutoRefine & 0.458 & 0.682 & 0.471 & 0.537 & 0.462 & 0.479 & 0.229 & 0.520 & 0.423 & 0.472 \\
            & Ours       & \textbf{0.496} & \textbf{0.687} & \textbf{0.477} & \textbf{0.553} & \textbf{0.494} & \textbf{0.508} & \textbf{0.265} & \textbf{0.528} & \textbf{0.449} & \textbf{0.494} \\
        \bottomrule
    \end{tabular}
    }
    \caption{Performance comparison (EM) across model families and scales. ``AutoRefine'' refers to the search-and-refine-during-think baseline, and ``Ours'' refers to our plan method. \textbf{Bold} indicates the higher value between  AutoRefine and our method.}
    \label{tab:main_results_em}
\end{table*}

%% file: tables/self_vs_distill.tex
\begin{table*}[!h]
\centering
\small
\resizebox{\linewidth}{!}{
\setlength{\tabcolsep}{4pt}
\begin{tabular}{llccccccccccc}
\toprule
& & \multicolumn{4}{c}{Single-Hop QA} & \multicolumn{5}{c}{Multi-Hop QA} & \\
\cmidrule(lr){3-6} \cmidrule(lr){7-11}
Model & Method & NQ & TriviaQA & PopQA & SH-Avg. & HotpotQA & 2Wiki & Musique & Bamb. & MH-Avg. & Avg. \\
\midrule
\multirow{2}{*}{Qwen2.5-3B-Base}
& Distillation (72B-Instruct) & 0.433 & 0.604 & 0.417 & 0.485 & 0.421 & 0.405 & 0.186 & \textbf{0.435} & 0.362 & 0.414 \\
& Self-Bootstrapping (Ours)   & \textbf{0.455} & \textbf{0.623} & \textbf{0.452} & \textbf{0.510} & \textbf{0.434} & \textbf{0.429} & \textbf{0.212} & 0.403 & \textbf{0.370} & \textbf{0.430} \\
\midrule
\multirow{2}{*}{Qwen2.5-7B-Base$^\dagger$}
& Distillation (72B-Instruct) & 0.448 & 0.658 & 0.450 & 0.519 & 0.455 & 0.446 & 0.216 & \textbf{0.484} & 0.400 & 0.451 \\
& Self-Bootstrapping (Ours)   & \textbf{0.490} & \textbf{0.665} & \textbf{0.468} & \textbf{0.541} & \textbf{0.464} & \textbf{0.447} & \textbf{0.227} & 0.464 & \textbf{0.401} & \textbf{0.461} \\
\midrule
\multirow{2}{*}{Qwen3-4B-Base$^\dagger$}
& Distillation (72B-Instruct) & 0.435 & 0.637 & 0.435 & 0.502 & 0.438 & 0.416 & 0.203 & \textbf{0.505} & 0.390 & 0.438 \\
& Self-Bootstrapping (Ours)   & \textbf{0.473} & \textbf{0.651} & \textbf{0.466} & \textbf{0.530} & \textbf{0.456} & \textbf{0.428} & \textbf{0.223} & 0.476 & \textbf{0.396} & \textbf{0.453} \\
\bottomrule
\end{tabular}}
\caption{Self-bootstrapping compared with \ strong-model distillation as the SFT data source, evaluated across three target models. Both variants use the same SFT-then-RL pipeline and 4{,}500 correctness-filtered trajectories, differing only in the data generator (Self-Bootstrapping Qwen2.5-3B-base vs.\ Qwen2.5-72B-Instruct). The evaluation metric is Exact Match; \textbf{bold} indicates the better result within each model group. $^\dagger$ means RL training with 72B-distilled SFT initialization collapsed in multiple runs for Qwen2.5-7B-Base and Qwen3-4B-Base; reported numbers are best checkpoints before collapse.}
\label{tab:self_vs_distill}
\end{table*}

%% file: tables/appendix_7b_rl.tex
\begin{table*}[!h]
\centering
\small
\resizebox{\linewidth}{!}{
\setlength{\tabcolsep}{4pt}
\begin{tabular}{lccccccccccc}
\toprule
& \multicolumn{4}{c}{Single-Hop QA} & \multicolumn{5}{c}{Multi-Hop QA} & \\
\cmidrule(lr){2-5} \cmidrule(lr){6-10}
Method & NQ & TriviaQA & PopQA & SH-Avg. & HotpotQA & 2Wiki & Musique & Bamb. & MH-Avg. & Avg. \\
\midrule
Direct RL (joint) & 0.445 & 0.620 & 0.430 & 0.498 & 0.408 & 0.389 & 0.173 & 0.368 & 0.335 & 0.405 \\
Self-Bootstrapping (Ours) & 0.490 & 0.665 & 0.468 & 0.541 & 0.464 & 0.447 & 0.227 & 0.464 & 0.401 & 0.461 \\
\midrule
$\Delta$ (Ours $-$ Direct) & \textbf{+0.045} & \textbf{+0.045} & \textbf{+0.038} & \textbf{+0.043} & \textbf{+0.056} & \textbf{+0.058} & \textbf{+0.054} & \textbf{+0.096} & \textbf{+0.066} & \textbf{+0.056} \\
\bottomrule
\end{tabular}}
\caption{Comparison of direct joint RL training and self-bootstrapping on Qwen2.5-7B-Base. Self-bootstrapping consistently outperforms direct RL despite the model satisfying the training feasibility conditions for joint acquisition. Numbers are Exact Match.}
\label{tab:7b_comparison}
\end{table*}

%% file: tables/sft_ablation.tex
\begin{table*}[!h]
\centering
\small
\resizebox{\linewidth}{!}{
\setlength{\tabcolsep}{4pt}
\begin{tabular}{lccccccccccc}
\toprule
& \multicolumn{4}{c}{Single-Hop QA} & \multicolumn{5}{c}{Multi-Hop QA} & \\
\cmidrule(lr){2-5} \cmidrule(lr){6-10}
\# SFT Traj. & NQ & TriviaQA & PopQA & SH-Avg. & HotpotQA & 2Wiki & Musique & Bamb. & MH-Avg. & Avg. \\
\midrule
1{,}500 & \textbf{0.455} & 0.621 & 0.448 & 0.508 & 0.426 & 0.427 & 0.205 & 0.382 & 0.360 & 0.423 \\
3{,}000 & 0.453 & \textbf{0.628} & 0.446 & 0.509 & 0.427 & 0.427 & 0.188 & \textbf{0.432} & 0.368 & 0.429 \\
4{,}500 (Default) & \textbf{0.455} & {0.623} & \textbf{0.452} & \textbf{0.510} & \textbf{0.434} & \textbf{0.429} & \textbf{0.212} & 0.403 & \textbf{0.370} & \textbf{0.430} \\
\bottomrule
\end{tabular}}
\caption{Effect of self-bootstrapped SFT data scale on final RL performance. All variants use the same 3B base model, SFT-then-RL pipeline, and RL recipe, differing only in the number of correctness-filtered seed-model trajectories used for SFT. Numbers are Exact Match.}
\label{tab:sft_scale}
\end{table*}

%% file: tables/plan_threshold.tex
\begin{table*}[!h]
\centering
\small
\resizebox{\linewidth}{!}{
\setlength{\tabcolsep}{4pt}
\begin{tabular}{lccccccccccc}
\toprule
& \multicolumn{4}{c}{Single-Hop QA} & \multicolumn{5}{c}{Multi-Hop QA} & \\
\cmidrule(lr){2-5} \cmidrule(lr){6-10}
Plan Threshold $\delta$ & NQ & TriviaQA & PopQA & SH-Avg. & HotpotQA & 2Wiki & Musique & Bamb. & MH-Avg. & Avg. \\
\midrule
Default (0.25) & \textbf{0.455} & \textbf{0.623} & \textbf{0.452} & \textbf{0.510} & \textbf{0.434} & \textbf{0.429} & \textbf{0.212} & 0.403 & \textbf{0.370} & \textbf{0.430} \\
$0.5$ & 0.448 & 0.606 & 0.430 & 0.495 & 0.412 & 0.415 & 0.187 & \textbf{0.444} & 0.365 & 0.420 \\
$0.75$ & 0.448 & 0.613 & 0.445 & 0.502 & 0.419 & 0.405 & 0.197 & 0.384 & 0.352 & 0.416 \\
\bottomrule
\end{tabular}}
\caption{Ablation on the plan-triggering threshold. We compare our default threshold against fixed thresholds of $0.5$ and $0.75$, keeping the base model, SFT-then-RL pipeline, and all other components identical.}
\label{tab:plan_threshold}
\end{table*}

%% file: tables/success_trajectory.tex
\begin{table*}[t]
\centering
\small
\begin{tabular}{p{0.95\textwidth}}
\toprule
\textbf{Question:} Who designed and built the ship that was used by the team leader that first crossed the Greenland interior to explore the Arctic and Antarctic? \\
\textbf{Ground Truth:} Colin Archer \\
\midrule
\texttt{<plan>} \\
\quad - Identify the team leader who first crossed the Greenland interior to explore the Arctic and Antarctic. \\
\quad - Determine the name of the ship used by this team leader. \\
\quad - Find out who designed and built this ship. \\
\texttt{</plan>} \\
\midrule
\textbf{Step 1.} \\
\texttt{<think>} First, I need to identify the team leader who first crossed the Greenland interior to explore the Arctic and Antarctic. \texttt{</think>} \\
\texttt{<search>} Who crossed the Greenland interior to explore the Arctic and Antarctic first? \texttt{</search>} \\
\texttt{<documents>} \ldots \textcolor{blue}{Fridtjof Nansen} \ldots led the team that made the first crossing of the Greenland interior in 1888 \ldots \texttt{</documents>} \\
\texttt{<refine>} The team leader who first crossed the Greenland interior is \textcolor{blue}{Fridtjof Nansen}. Now I need to determine the name of the ship used by him. \texttt{</refine>} \\
\midrule
\textbf{Step 2.} \\
\texttt{<think>} Next, I need to determine the name of the ship used by Fridtjof Nansen. \texttt{</think>} \\
\texttt{<search>} Which ship was used by Fridtjof Nansen during his first crossing of the Greenland interior? \texttt{</search>} \\
\texttt{<documents>} \ldots \textcolor{blue}{Fram} \ldots is a ship that was used in expeditions of the Arctic and Antarctic regions by the Norwegian explorers Fridtjof Nansen \ldots \texttt{</documents>} \\
\texttt{<refine>} The ship used by Fridtjof Nansen is the \textcolor{blue}{Fram}. \texttt{</refine>} \\
\midrule
\textbf{Step 3.} \\
\texttt{<think>} Finally, I need to find out who designed and built the Fram. \texttt{</think>} \\
\texttt{<search>} Who designed and built the ship Fram? \texttt{</search>} \\
\texttt{<documents>} \ldots Fram was designed and built by the Scottish-Norwegian shipwright \textcolor{blue}{Colin Archer} for Fridtjof Nansen's 1893 Arctic expedition \ldots \texttt{</documents>} \\
\texttt{<refine>} The Fram was designed and built by \textcolor{blue}{Colin Archer}. \texttt{</refine>} \\
\midrule
\texttt{<answer>} Colin Archer \texttt{</answer>} \\
\bottomrule
\end{tabular}
\caption{A representative trajectory generated by our method, illustrating plan-based multi-hop reasoning. Key evidence is highlighted in \textcolor{blue}{blue}.}
\label{tab:trajectory_case}
\end{table*}

%% file: tables/case_freud.tex
\begin{table*}[t]
\centering
\small
\renewcommand{\arraystretch}{1.25}
\setlength{\tabcolsep}{6pt}
\begin{tabularx}{\textwidth}{@{}X@{\hspace{1.2em}}X@{}}
\toprule
\multicolumn{2}{@{}l}{\textbf{Question:} \emph{For which club did the winner of the 2007 Ballon d'Or play in 2012?}\quad\textbf{Ground Truth:} Real Madrid}\\
\midrule
\textbf{Reactive Baseline} \hfill \xmark~\emph{Lionel Messi} &
\textbf{Ours (Plan)} \hfill \cmark~\emph{Real Madrid} \\
\midrule
\textit{(no plan generated)} &
\texttt{<plan>}\newline
\hspace*{1em}- Identify the winner of the 2007 Ballon d'Or.\newline
\hspace*{1em}- Determine the club that winner played for in 2012.\newline
\texttt{</plan>} \\
\midrule
\multicolumn{2}{@{}l}{\textbf{Step 1.}} \\
\addlinespace[-2pt]
\drifttag~\texttt{<search>} winner of the 2007 Ballon d'Or and 2012 club \texttt{</search>}\newline
\texttt{<docs>} \ldots \ev{2011 FIFA Ballon d'Or} \ldots Lionel Messi \ldots \ev{2012 FIFA Ballon d'Or} \ldots Lionel Messi \ldots a record fourth title \ldots \texttt{</docs>}\newline
\texttt{<refine>} \leakt{The winner is Lionel Messi.} \emph{(query conflates two sub-questions; retrieves off-year award pages)} \texttt{</refine>}
&
\texttt{<think>} First, I need to find the winner of the 2007 Ballon d'Or. \texttt{</think>}\newline
\texttt{<search>} Who won the 2007 Ballon d'Or? \texttt{</search>}\newline
\texttt{<docs>} \ldots the 2007 Ballon d'Or was awarded to \ev{Kaká} \ldots \texttt{</docs>}\newline
\texttt{<refine>} The winner of the 2007 Ballon d'Or was \ev{Kaká}. \texttt{</refine>} \\
\midrule
\texttt{<answer>} \leakt{Lionel Messi} \texttt{</answer>}~\xmark
&
\multicolumn{1}{@{}l@{}}{\textbf{Step 2.}}\\
\addlinespace[-2pt]
&
\texttt{<think>} Now I need to find which club Kaká played for in 2012. \texttt{</think>}\newline
\texttt{<search>} Which club did Kaká play for in 2012? \texttt{</search>}\newline
\texttt{<docs>} \ldots Kaká \ldots at \ev{Real Madrid} \ldots finished the 2011--12 season in second place in La Liga \ldots \texttt{</docs>}\newline
\texttt{<refine>} Kaká played for \ev{Real Madrid} in 2012. \texttt{</refine>}\newline
\texttt{<answer>} \ev{Real Madrid} \texttt{</answer>}~\cmark \\
\midrule
\multicolumn{2}{@{}p{\dimexpr\textwidth-2\tabcolsep}@{}}{\textbf{Analysis.}~Lacking a decomposition step, the reactive baseline collapses the two-hop question into a single query that mixes the \emph{2007 award} with the \emph{2012 club}. The query no longer targets either sub-question cleanly, and retrieval drifts to adjacent award years (2011/2012), binding the answer to that year's winner, Messi. The plan-based model fixes the two hops as ordered sub-goals before retrieval: Step~1 queries only the 2007 winner (\emph{Kaká}), and Step~2's query is anchored to that resolved entity, retrieving Kaká's 2012 club. Decomposition keeps each query tied to a pre-specified sub-question, preventing the cross-hop query contamination that drives the baseline off-target.} \\
\bottomrule
\end{tabularx}
\caption{
    \textbf{Case: Cross-hop query drift.} Without a plan, the agent fuses both hops into one query and retrieves the wrong award year, answering with Messi. The plan anchors each search to a pre-specified sub-question, yielding the correct club. Each Plan step follows the full \texttt{think}$\rightarrow$\texttt{search}$\rightarrow$\texttt{refine} format. Key evidence is highlighted in \ev{blue}; the drifted reasoning that produces the wrong answer in \leakt{red}.
}
\label{tab:case_ballondor}
\end{table*}

%% file: tables/fail_trajectory.tex
\begin{table*}[t]
\centering
\small
\begin{tabular}{p{0.95\textwidth}}
\toprule

\rowcolor{gray!20}
\textbf{Case 1 — Prior collapse on instruction-tuned models.} The trajectory preserves the surface format but bypasses all reasoning: \texttt{<think>} blocks are missing, search queries are duplicated, and the refine step hallucinates an answer absent from the retrieved documents. \\
\midrule
\textbf{Question:} Who sings ``All I Can Think About Is Getting You Home''? \\
\textbf{Ground Truth:} Chris Young \quad \textbf{Model Output:} \textcolor{red}{Guy Sebastian} \\[2pt]
\texttt{<plan>} \\
\quad - Step 1: Find who sings ``All I Can Think About Is Getting You Home'' \\
\quad - Step 2: Identify the singer \\
\texttt{</plan>} \\[2pt]
\textcolor{red}{[\texttt{<think>} block missing]} \\
\texttt{<search>}Who sings All I Can Think About Is Getting You Home\texttt{</search>} \\
\texttt{<documents>} Doc 1 (``Gettin' You Home (The Black Dress Song)''): ...songwriter Cory Batten suggested the lyric ``all I can think about is gettin' you home''... \quad Doc 2 (``All I Can Think About Is You'' -- Coldplay)... \quad Doc 3 (``All I Do Is Think of You'' -- The Supremes)... \texttt{</documents>} \\
\texttt{<refine>}From the documents retrieved, the song is sung by \textcolor{red}{Guy Sebastian}. I will now search for more information.\texttt{</refine>} \\[2pt]
\textcolor{red}{[\texttt{<think>} missing; identical query reissued]} \\
\texttt{<search>}Who sings All I Can Think About Is Getting You Home\texttt{</search>} \\
\texttt{<documents>} [identical documents returned] \texttt{</documents>} \\
\texttt{<refine>}From the documents, the singer is \textcolor{red}{Guy Sebastian}.\texttt{</refine>} \\
\texttt{<answer>}\textcolor{red}{Guy Sebastian}\texttt{</answer>} \\[4pt]
\midrule
\rowcolor{gray!20}
\textbf{Case 2 — Structural and lexical breakdown on Qwen3.} The \texttt{<think>} blocks are empty or misplaced, and the output ultimately degenerates into incoherent symbol fragments. This is consistent with the late-stage entropy explosion observed in the training dynamics. \\
\midrule
\textbf{Question:} What American actress and comedian studied alongside Doreen Cannon? \\
\textbf{Ground Truth:} Anne Meara \quad \textbf{Model Output:} \textcolor{red}{Dyan Cannon} (followed by lexical breakdown) \\[2pt]
\texttt{<think>}\textcolor{red}{[empty]}\texttt{</think>} \\
\texttt{<plan>} \\\quad \textit{(illegal marker inside plan)} \\
\quad Step 1: Identify the person who studied alongside Doreen Cannon. \\
\quad Step 2: Determine if the person is an American actress and comedian. \\
\texttt{</plan>} \\
\textcolor{red}{\texttt{</think>}} \quad \textit{(dangling closing tag with no matching opener)} \\[2pt]
Step 1: ... \quad \texttt{<search>}who studied alongside Doreen Cannon\texttt{</search>} \\
\texttt{<documents>} ...Her contemporaries and fellow students were Peter Falk, Geraldine Page, Sandy Dennis, Maureen Stapleton, \textcolor{blue}{Anne Meara} and Jerry Stiller... \texttt{</documents>} \\
\texttt{<refine>}Step 1: Identify the person who studied alongside Doreen Cannon.\texttt{</refine>} \quad \textit{(refine merely restates the sub-question; no evidence integrated)} \\[2pt]
\textit{Step 2 retrieves a passage about Dyan Cannon, a different person, and the model anchors on this distractor:} \\
``\textcolor{red}{Dyan Cannon} is an American actress and comedian who studied alongside Doreen Cannon.'' \\[2pt]
\textit{After several reissued invalid actions, the output collapses into incoherent fragments:} \\
\textcolor{red}{``Dyan Cannon is concluded matching accurate fitting \$ \textbackslash textnormal Actual resultant Actively thus \$ \$ Answer from Active Zone Spaced herein thus upto(\&\$ Actress\$ Key \$(\$ concluded. Accent \& thus \$(\$ Area \& the''} \\[4pt]
\midrule
\rowcolor{gray!20}
\textbf{Case 3 — Missing prerequisite sub-skills on Qwen2.5-3B-Base (P$\to$R).} The trajectory starts in a structurally valid form---a well-formed \texttt{<plan>}, a coherent \texttt{<think>}, and a focused \texttt{<search>}---but breaks down at the think stage: lacking prior refine training, the model emits malformed tags, repeatedly triggers invalid-action errors, and degenerates into off-topic loops without ever integrating the retrieved evidence. \\
\midrule
\textbf{Question:} When does Pam tell Jim how she feels? \\
\textbf{Ground Truth:} Beach Games \quad \textbf{Model Output:} \textcolor{red}{and} \\[2pt]
\texttt{<plan>} \\
\quad - Step 1: Identify when Pam tells Jim how she feels. \\
\quad - Step 2: Determine the specific moment or episode where this happens. \\
\texttt{</plan>} \\[2pt]
\texttt{<think>}I need to identify the specific moment in the series where Pam reveals her feelings to Jim.\texttt{</think>} \\
\texttt{<search>}When does Pam tell Jim how she feels in The Office\texttt{</search>} \\
\texttt{<documents>} Doc 1 (``Pam Beesly''): ...In ``Casino Night'' she turns him down... \quad Doc 2 (``Body Language''): ...Pam and Jim work on their first sales pitch together... \quad Doc 3 (``Paper Airplane''): ... \texttt{</documents>} \\[2pt]
\textcolor{red}{\texttt{<think]}Based on the document selected, I need to find which episode Pam tells Jim she feels feelings.\texttt{</thought></}} \quad \textit{(malformed tags where a refine step should appear)} \\
\textit{System:} ``My previous action is invalid... Let me try again.'' \\[2pt]
\textit{The model reissues invalid actions and drifts off-topic instead of refining the evidence:} \\
\textcolor{red}{\texttt{<think]}I realized I went wrong earlier; for such requirements chatGPT originally belongs to GPT; however, I am ending the activity within docs.\texttt{</thought></}} \\
\textit{(several more invalid actions follow; no \texttt{<refine>} is ever produced, and the run terminates with the extracted answer)} \textcolor{red}{``and''} \\[4pt]

\bottomrule
\end{tabular}
\caption{Representative failure trajectories under direct RL training on the plan task. }
\label{tab:failure_cases}
\end{table*}

%% file: tables/sizes_and_types_appendix.tex
\setlength{\LTleft}{0pt}
\setlength{\LTright}{0pt plus 1fil}

\footnotesize
\setlength{\tabcolsep}{1.5pt}
\renewcommand{\arraystretch}{0.95}

\begin{longtable}{lllcccccccccc}
    \caption{Detailed performance comparison (EM, F1, CEM) across model families and scales. ``AutoRefine'' refers to the search-and-refine-during-think baseline, and ``Ours'' refers to our plan-and-refine method. For Qwen2.5-3B-Instruct, F1 and CEM are not reported for AutoRefine (shown as ``---'') because the original paper~\citep{shi2025search} only reported EM for this configuration.}
    \label{tab:detailed_results} \\
    \toprule
    & & & \multicolumn{4}{c}{Single-Hop QA} & \multicolumn{5}{c}{Multi-Hop QA} & \\
    \cmidrule(lr){4-7} \cmidrule(lr){8-12}
    Model & Method & Metric & NQ & TriviaQA & PopQA & SH-Avg. & HotpotQA & 2Wiki & Musique & Bamb. & MH-Avg. & Avg. \\
    \midrule
    \endfirsthead
    
    \multicolumn{13}{c}{\tablename\ \thetable\ -- \textit{Continued from previous page}} \\
    \toprule
    & & & \multicolumn{4}{c}{Single-Hop QA} & \multicolumn{5}{c}{Multi-Hop QA} & \\
    \cmidrule(lr){4-7} \cmidrule(lr){8-12}
    Model & Method & Metric & NQ & TriviaQA & PopQA & SH-Avg. & HotpotQA & 2Wiki & Musique & Bamb. & MH-Avg. & Avg. \\
    \midrule
    \endhead
    
    \midrule
    \multicolumn{13}{r}{\textit{Continued on next page}} \\
    \endfoot
    
    \bottomrule
    \endlastfoot
    
    \rowcolor[gray]{0.9}
    \multicolumn{13}{l}{\textit{Qwen2.5 Family}} \\
    \multirow{6}{*}{Qwen2.5-3B-Base}
        & \multirow{3}{*}{AutoRefine}
          & EM  & 0.467 & 0.620 & 0.450 & 0.512 & 0.405 & 0.393 & 0.157 & 0.344 & 0.325 & 0.405 \\
        & & F1  & 0.534 & 0.689 & 0.479 & 0.567 & 0.503 & 0.453 & 0.233 & 0.449 & 0.410 & 0.477 \\
        & & CEM & 0.502 & 0.674 & 0.468 & 0.548 & 0.440 & 0.428 & 0.175 & 0.384 & 0.357 & 0.439 \\
        \cmidrule(lr){2-13}
        & \multirow{3}{*}{Ours}
          & EM  & 0.455 & 0.623 & 0.452 & 0.510 & 0.434 & 0.429 & 0.212 & 0.403 & 0.370 & 0.430 \\
        & & F1  & 0.541 & 0.698 & 0.497 & 0.579 & 0.552 & 0.488 & 0.304 & 0.531 & 0.469 & 0.516 \\
        & & CEM & 0.494 & 0.676 & 0.474 & 0.548 & 0.473 & 0.471 & 0.238 & 0.435 & 0.404 & 0.466 \\
    \midrule
    \multirow{6}{*}{Qwen2.5-3B-Instruct}
        & \multirow{3}{*}{AutoRefine}
          & EM  & 0.436 & 0.597 & 0.447 & 0.493 & 0.404 & 0.380 & 0.169 & 0.336 & 0.322 & 0.396 \\
        & & F1  & --- & --- & --- & --- & --- & --- & --- & --- & --- & --- \\
        & & CEM & --- & --- & --- & --- & --- & --- & --- & --- & --- & --- \\
        \cmidrule(lr){2-13}
        & \multirow{3}{*}{Ours}
          & EM  & 0.462 & 0.631 & 0.460 & 0.518 & 0.437 & 0.422 & 0.207 & 0.408 & 0.369 & 0.432 \\
        & & F1  & 0.549 & 0.708 & 0.503 & 0.587 & 0.559 & 0.484 & 0.301 & 0.532 & 0.469 & 0.519 \\
        & & CEM & 0.504 & 0.689 & 0.482 & 0.558 & 0.481 & 0.470 & 0.239 & 0.432 & 0.405 & 0.471 \\
    \midrule
    \multirow{6}{*}{Qwen2.5-7B-Base}
        & \multirow{3}{*}{AutoRefine}
          & EM  & 0.484 & 0.659 & 0.487 & 0.543 & 0.451 & 0.405 & 0.187 & 0.512 & 0.389 & 0.455 \\
        & & F1  & 0.574 & 0.729 & 0.525 & 0.609 & 0.573 & 0.467 & 0.283 & 0.604 & 0.482 & 0.536 \\
        & & CEM & 0.523 & 0.707 & 0.500 & 0.577 & 0.487 & 0.441 & 0.217 & 0.528 & 0.418 & 0.486 \\
        \cmidrule(lr){2-13}
        & \multirow{3}{*}{Ours}
          & EM  & 0.490 & 0.665 & 0.468 & 0.541 & 0.464 & 0.447 & 0.227 & 0.464 & 0.401 & 0.461 \\
        & & F1  & 0.580 & 0.737 & 0.512 & 0.610 & 0.588 & 0.519 & 0.329 & 0.601 & 0.509 & 0.552 \\
        & & CEM & 0.527 & 0.714 & 0.486 & 0.576 & 0.503 & 0.498 & 0.267 & 0.488 & 0.439 & 0.498 \\
    \midrule
\multirow{6}{*}{Qwen2.5-7B-Instruct}
    & \multirow{3}{*}{AutoRefine}
      & EM  & 0.423 & 0.627 & 0.430 & 0.493 & 0.409 & 0.348 & 0.175 & 0.440 & 0.343 & 0.407 \\
    & & F1  & 0.519 & 0.704 & 0.472 & 0.565 & 0.469 & 0.469 & 0.256 & 0.527 & 0.430 & 0.488 \\
    & & CEM & 0.480 & 0.684 & 0.452 & 0.539 & 0.443 & 0.383 & 0.198 & 0.456 & 0.370 & 0.442 \\
        \cmidrule(lr){2-13}
        & \multirow{3}{*}{Ours}
          & EM  & 0.467 & 0.641 & 0.461 & 0.523 & 0.452 & 0.447 & 0.231 & 0.468 & 0.400 & 0.452 \\
        & & F1  & 0.558 & 0.720 & 0.507 & 0.595 & 0.574 & 0.516 & 0.331 & 0.574 & 0.499 & 0.540 \\
        & & CEM & 0.513 & 0.700 & 0.483 & 0.566 & 0.494 & 0.496 & 0.270 & 0.500 & 0.440 & 0.494 \\
    \midrule
    \multirow{6}{*}{Qwen2.5-14B-Base}
        & \multirow{3}{*}{AutoRefine}
          & EM  & 0.516 & 0.693 & 0.489 & 0.566 & 0.468 & 0.468 & 0.230 & 0.536 & 0.426 & 0.486 \\
        & & F1  & 0.604 & 0.773 & 0.534 & 0.637 & 0.630 & 0.543 & 0.328 & 0.639 & 0.535 & 0.579 \\
        & & CEM & 0.563 & 0.755 & 0.516 & 0.612 & 0.549 & 0.533 & 0.268 & 0.560 & 0.478 & 0.535 \\
        \cmidrule(lr){2-13}
        & \multirow{3}{*}{Ours}
          & EM  & 0.514 & 0.694 & 0.487 & 0.565 & 0.500 & 0.484 & 0.269 & 0.512 & 0.441 & 0.494 \\
        & & F1  & 0.606 & 0.769 & 0.533 & 0.636 & 0.630 & 0.562 & 0.375 & 0.647 & 0.554 & 0.589 \\
        & & CEM & 0.555 & 0.748 & 0.511 & 0.605 & 0.547 & 0.551 & 0.311 & 0.537 & 0.486 & 0.537 \\
    \midrule
    \multirow{6}{*}{Qwen2.5-14B-Instruct}
        & \multirow{3}{*}{AutoRefine}
          & EM  & 0.483 & 0.673 & 0.451 & 0.535 & 0.463 & 0.423 & 0.231 & 0.512 & 0.407 & 0.462 \\
        & & F1  & 0.574 & 0.747 & 0.495 & 0.605 & 0.584 & 0.492 & 0.321 & 0.628 & 0.506 & 0.549 \\
        & & CEM & 0.532 & 0.727 & 0.478 & 0.579 & 0.505 & 0.482 & 0.262 & 0.545 & 0.449 & 0.504 \\
        \cmidrule(lr){2-13}
        & \multirow{3}{*}{Ours}
          & EM  & 0.510 & 0.694 & 0.483 & 0.563 & 0.499 & 0.493 & 0.260 & 0.528 & 0.445 & 0.496 \\
        & & F1  & 0.600 & 0.771 & 0.529 & 0.634 & 0.627 & 0.564 & 0.367 & 0.686 & 0.561 & 0.592 \\
        & & CEM & 0.553 & 0.750 & 0.507 & 0.603 & 0.554 & 0.554 & 0.303 & 0.561 & 0.493 & 0.540 \\
    \midrule
    \rowcolor[gray]{0.9}
    \multicolumn{13}{l}{\textit{Llama3.2 Family}} \\
    \multirow{6}{*}{Llama3.2-3B-Base}
        & \multirow{3}{*}{AutoRefine}
          & EM  & 0.267 & 0.488 & 0.280 & 0.345 & 0.239 & 0.248 & 0.030 & 0.087 & 0.151 & 0.234 \\
        & & F1  & 0.341 & 0.556 & 0.328 & 0.408 & 0.315 & 0.321 & 0.127 & 0.179 & 0.236 & 0.310 \\
        & & CEM & 0.289 & 0.512 & 0.288 & 0.363 & 0.243 & 0.271 & 0.046 & 0.098 & 0.165 & 0.250 \\
        \cmidrule(lr){2-13}
        & \multirow{3}{*}{Ours}
          & EM  & 0.443 & 0.612 & 0.450 & 0.502 & 0.411 & 0.388 & 0.180 & 0.368 & 0.337 & 0.408 \\
        & & F1  & 0.533 & 0.688 & 0.494 & 0.572 & 0.529 & 0.461 & 0.268 & 0.476 & 0.434 & 0.493 \\
        & & CEM & 0.489 & 0.668 & 0.471 & 0.543 & 0.451 & 0.435 & 0.214 & 0.384 & 0.371 & 0.445 \\
    \midrule
    \multirow{6}{*}{Llama3.2-3B-Instruct}
        & \multirow{3}{*}{AutoRefine}
          & EM  & 0.437 & 0.638 & 0.419 & 0.498 & 0.375 & 0.225 & 0.169 & 0.349 & 0.280 & 0.373 \\
        & & F1  & 0.543 & 0.705 & 0.458 & 0.569 & 0.475 & 0.252 & 0.237 & 0.462 & 0.357 & 0.447 \\
        & & CEM & 0.477 & 0.688 & 0.428 & 0.531 & 0.398 & 0.261 & 0.182 & 0.374 & 0.304 & 0.401 \\
        \cmidrule(lr){2-13}
        & \multirow{3}{*}{Ours}
          & EM  & 0.478 & 0.626 & 0.459 & 0.521 & 0.443 & 0.390 & 0.197 & 0.468 & 0.375 & 0.437 \\
        & & F1  & 0.562 & 0.701 & 0.501 & 0.588 & 0.563 & 0.450 & 0.287 & 0.583 & 0.471 & 0.521 \\
        & & CEM & 0.516 & 0.677 & 0.481 & 0.558 & 0.487 & 0.434 & 0.233 & 0.492 & 0.412 & 0.474 \\
    \midrule
    \rowcolor[gray]{0.9}
    \multicolumn{13}{l}{\textit{Qwen3 Family}} \\
    \multirow{6}{*}{Qwen3-4B-Base}
        & \multirow{3}{*}{AutoRefine}
          & EM  & 0.418 & 0.642 & 0.500 & 0.520 & 0.418 & 0.361 & 0.157 & 0.464 & 0.350 & 0.423 \\
        & & F1  & 0.513 & 0.707 & 0.579 & 0.600 & 0.503 & 0.438 & 0.250 & 0.554 & 0.436 & 0.506 \\
        & & CEM & 0.495 & 0.699 & 0.579 & 0.591 & 0.442 & 0.416 & 0.204 & 0.476 & 0.385 & 0.473 \\
        \cmidrule(lr){2-13}
        & \multirow{3}{*}{Ours}
          & EM  & 0.473 & 0.651 & 0.466 & 0.530 & 0.456 & 0.428 & 0.223 & 0.476 & 0.396 & 0.453 \\
        & & F1  & 0.563 & 0.728 & 0.513 & 0.601 & 0.577 & 0.499 & 0.321 & 0.599 & 0.499 & 0.543 \\
        & & CEM & 0.520 & 0.707 & 0.493 & 0.573 & 0.498 & 0.482 & 0.262 & 0.500 & 0.436 & 0.495 \\
    \midrule
    \multirow{6}{*}{Qwen3-4B-Instruct}
        & \multirow{3}{*}{AutoRefine}
          & EM  & 0.435 & 0.633 & 0.443 & 0.504 & 0.398 & 0.368 & 0.205 & 0.460 & 0.358 & 0.420 \\
        & & F1  & 0.538 & 0.723 & 0.492 & 0.584 & 0.505 & 0.421 & 0.278 & 0.549 & 0.438 & 0.501 \\
        & & CEM & 0.484 & 0.708 & 0.479 & 0.557 & 0.425 & 0.401 & 0.229 & 0.476 & 0.383 & 0.458 \\
        \cmidrule(lr){2-13}
        & \multirow{3}{*}{Ours}
          & EM  & 0.437 & 0.624 & 0.448 & 0.503 & 0.437 & 0.448 & 0.217 & 0.488 & 0.398 & 0.443 \\
        & & F1  & 0.535 & 0.703 & 0.496 & 0.578 & 0.556 & 0.522 & 0.310 & 0.603 & 0.498 & 0.534 \\
        & & CEM & 0.502 & 0.679 & 0.481 & 0.554 & 0.480 & 0.506 & 0.254 & 0.520 & 0.440 & 0.489 \\
    \midrule
    \multirow{6}{*}{Qwen3-8B}
        & \multirow{3}{*}{AutoRefine}
          & EM  & 0.450 & 0.665 & 0.470 & 0.528 & 0.421 & 0.369 & 0.219 & 0.496 & 0.376 & 0.441 \\
        & & F1  & 0.542 & 0.732 & 0.508 & 0.594 & 0.539 & 0.423 & 0.299 & 0.614 & 0.469 & 0.522 \\
        & & CEM & 0.471 & 0.700 & 0.479 & 0.550 & 0.440 & 0.389 & 0.239 & 0.504 & 0.393 & 0.460 \\
        \cmidrule(lr){2-13}
        & \multirow{3}{*}{Ours}
          & EM  & 0.454 & 0.660 & 0.476 & 0.530 & 0.461 & 0.453 & 0.233 & 0.475 & 0.406 & 0.459 \\
        & & F1  & 0.547 & 0.735 & 0.518 & 0.600 & 0.583 & 0.521 & 0.328 & 0.587 & 0.505 & 0.546 \\
        & & CEM & 0.500 & 0.711 & 0.496 & 0.569 & 0.497 & 0.501 & 0.274 & 0.484 & 0.439 & 0.495 \\
    \midrule
    \multirow{6}{*}{Qwen3-14B}
        & \multirow{3}{*}{AutoRefine}
          & EM  & 0.458 & 0.682 & 0.471 & 0.537 & 0.462 & 0.479 & 0.229 & 0.520 & 0.423 & 0.472 \\
        & & F1  & 0.562 & 0.759 & 0.532 & 0.618 & 0.580 & 0.542 & 0.316 & 0.632 & 0.517 & 0.560 \\
        & & CEM & 0.536 & 0.737 & 0.529 & 0.601 & 0.497 & 0.527 & 0.257 & 0.545 & 0.456 & 0.518 \\
        \cmidrule(lr){2-13}
        & \multirow{3}{*}{Ours}
          & EM  & 0.496 & 0.687 & 0.477 & 0.553 & 0.494 & 0.508 & 0.265 & 0.528 & 0.449 & 0.494 \\
        & & F1  & 0.587 & 0.764 & 0.526 & 0.626 & 0.620 & 0.578 & 0.364 & 0.677 & 0.560 & 0.588 \\
        & & CEM & 0.546 & 0.743 & 0.506 & 0.598 & 0.536 & 0.566 & 0.305 & 0.568 & 0.494 & 0.539 \\
\end{longtable}

\twocolumn  